\documentclass[12pt,twoside]{article}   
\usepackage[super,sort,comma]{natbib}

\usepackage{fancyhdr}		

\usepackage[section]{placeins}   %

\usepackage{graphicx}

\makeatletter \renewcommand\@biblabel[1]{$^{#1}$} \makeatother
\newcommand{\note}[1]{\mbox{}\\ \noindent \rule{16cm}{0.5mm} \\
{\em #1} \\ \noindent \rule{16cm}{0.5mm}
\typeout{    }
\typeout{***********note active on this page *************************}
\typeout{Note: #1  }
\typeout{****************************************end Note}
}

\newcommand{\cen}[1]{\begin{center} #1 \end{center}}

\usepackage[mathlines]{lineno}

\usepackage{hyperref}
\hypersetup{ colorlinks,
    citecolor=blue,
    filecolor=blue,
    linkcolor=blue,
    urlcolor=blue
}

\usepackage{xcolor}

\definecolor{gray}{rgb}{0.6,0.6,0.6}
\definecolor{red}{rgb}{0.85,0,0}
\definecolor{green}{rgb}{0,0.85,0}
\definecolor{blue}{rgb}{0,0,0.85}
\definecolor{beige}{rgb}{0.92,0.87,0.78}
\usepackage[all]{hypcap}    

\usepackage{amsmath,amssymb,amsfonts}
\usepackage[caption=false]{subfig}
\usepackage{algorithm,algorithmic}
\usepackage{textcomp}
\usepackage{multirow}
\usepackage{siunitx}
\usepackage{arydshln}

\usepackage{cleveref}
\crefname{equation}{Eq.}{Eqs.}
\crefname{figure}{Fig.}{Figs.}
\crefname{subfigure}{Fig.}{Figs.}
\crefname{table}{Table}{Tables}
\crefname{appendix}{Appendix}{Appendices}
\crefname{section}{Sec.}{Secs.}

\begin{document}

\cen{\sf {\Large {\bfseries Efficient Noise Context-Aware Representation for Low-Dose CT Denoising}} \\  
\vspace*{10mm}
{\large Minwoo Yu$^{1}$, N. Robert Bennett$^{2}$, Jongduk Baek$^{1,*}$, and Adam S. Wang$^{2,3,*}$} \\
$^{1}$Department of Artificial Intelligence, Yonsei University, Seoul, South Korea \\
$^{2}$Department of Radiology, Stanford University, Stanford, CA 94305, USA \\
$^{3}$Department of Electrical Engineering, Stanford University, Stanford, CA 94305, USA
\vspace{5mm}\\
}

\pagenumbering{roman}
\setcounter{page}{1}
\pagestyle{plain}

\begin{abstract}
\noindent {\bf Background:} While deep learning-based denoising has become widely adopted in low-dose CT, conventional models use generic architectures designed for natural images, do not explicitly account for non-stationary and spatially correlated CT noise characteristics.\\ 
{\bf Purpose:} To address this, we propose an Efficient Noise COntext-aware REpresentation (ENCORE) framework that explicitly leverages CT noise characteristics and anatomical features.\\
{\bf Methods:} First, we reformulate the noise synthesis procedure based on a realistic noise distribution beyond the conventional Gaussian approximation, providing a statistically motivated formulation for training pair generation. 
Next, we extract local noise contexts to guide the denoising process.
To fully leverage the potential of noise context, we propose a FlyingConv module, which adaptively changes convolution weights for each local image region. 
Furthermore, manipulating the intensity of the noise context maps at inference time enables zero-shot conditional denoising, allowing for dynamic control over the output image texture.\\
{\bf Results:} Rather than relying solely on increasing model parameters, our approach demonstrates that explicitly accounting for noise context can improve the quality-efficiency trade-off across simulated and real-world datasets, achieving higher fidelity with reduced computational complexity.\\
{\bf Conclusions:} ENCORE provides an effective and flexible paradigm for CT image denoising by bridging statistical noise modeling, architectural weight adaptation, and zero-shot clinical control. 
The entire codebase is available at \href{https://github.com/minwoo-yu/ENCORE.git}{https://github.com/minwoo-yu/ENCORE.git}.\\
\end{abstract}

\note{Keywords: Convolution with Adaptive Weight, CT image denoising, Noise Autocorrelation Contexts}

\newpage     


\newpage

\setlength{\baselineskip}{0.7cm}      

\pagenumbering{arabic}
\setcounter{page}{1}
\pagestyle{fancy}
\section{Introduction}
\label{sec:introduction}
As deep learning (DL)-based denoising has become widely adopted in low-dose CT imaging \cite{chen2017low}, extensive research has focused on building training frameworks that reflect CT-specific characteristics.
This involves either enhancing perceptual quality through specialized loss functions \cite{yang2018low, wolterink2017generative, han2021low,kim2019performance, zhao2025nifa} or exploring self-supervised learning pipelines for clinical settings where paired data are difficult to obtain \cite{noise2noise, half2half, CTnoise2noise, kim2021weakly}.
However, from an architectural perspective, relatively few attempts have been made to design or modify denoising networks tailored for CT imaging.
Fundamentally, the raw projection data exhibit a mixed noise structure: Poisson quantum noise stemming from photon-counting statistics and Gaussian electronic noise introduced by system components.
In addition, during reconstruction, the backprojection operation induces complex spatial correlations, thereby modulating and amplifying the noise spectrum in the reconstructed image domain.
In practice, standard denoising architectures typically take only a single noisy image as input and perform best under the assumption of independent and identically distributed (i.i.d.) noise \cite{plotz2017benchmarking}. 
As a result, their performance is limited on CT images, where this i.i.d. assumption is violated due to spatial correlation and scan-dependent variability.
This underscores the critical need for a model-side approach capable of directly exploiting CT noise properties.

One notable attempt to exploit unique CT noise properties is the noise-augmented deep denoising (NADD) approach \cite{NADD}. 
Rather than relying solely on the low-dose (LD) image, NADD incorporates synthesized noise maps to better capture non-stationary CT noise characteristics.
Nonetheless, directly providing stochastic noise realizations may make it difficult for the network to extract noise information. 
Therefore, simply concatenating stochastic noise maps without any proper preprocessing to stabilize the fluctuations may introduce unwanted variance, potentially making it challenging for the network to accurately capture CT noise characteristics.
Furthermore, from a structural perspective, NADD primarily utilizes input channel expansion, suggesting that architectural modifications could further improve denoising performance.

Accordingly, we propose an Efficient Noise COntext-aware REpresentation (ENCORE) structure for CT image denoising, in which the model adaptively exploits the noise characteristics of the given data to enhance both restoration performance and computational efficiency.
While our ENCORE addresses the clinical scarcity of paired training data by adopting a Noise2Noise (N2N)-based self-supervised learning pipeline, it focuses on enhancing statistical fidelity of the synthesized noise. 
In specific, rather than relying on a standard Gaussian approximation \cite{CTnoise2noise}, we reformulate the noise synthesis procedure based on the Cornish-Fisher expansion \cite{cornish1938moments} to achieve more accurate CT noise simulation.
Subsequently, to provide the model with more direct and stable context guidance than NADD's raw stochastic noise augmentations \cite{NADD}, we introduce an efficient autocovariance estimation step to explicitly extract local noise autocovariance.
In addition, our framework natively supports zero-shot conditional denoising, allowing the model to dynamically balance noise reduction and texture preservation during inference by simply scaling the context maps. 
To mitigate the latency overhead in noise augmentation and context estimation, we customize the reconstruction kernels to enhance throughput compared to conventional open-source FBP algorithms.
Finally, to embed the noise context directly within the model, we redesign the network layers by introducing an on-the-fly weight-modulated convolution (FlyingConv) operation. 
 
\begin{figure}
    \centering
        \includegraphics[width=0.9\linewidth]{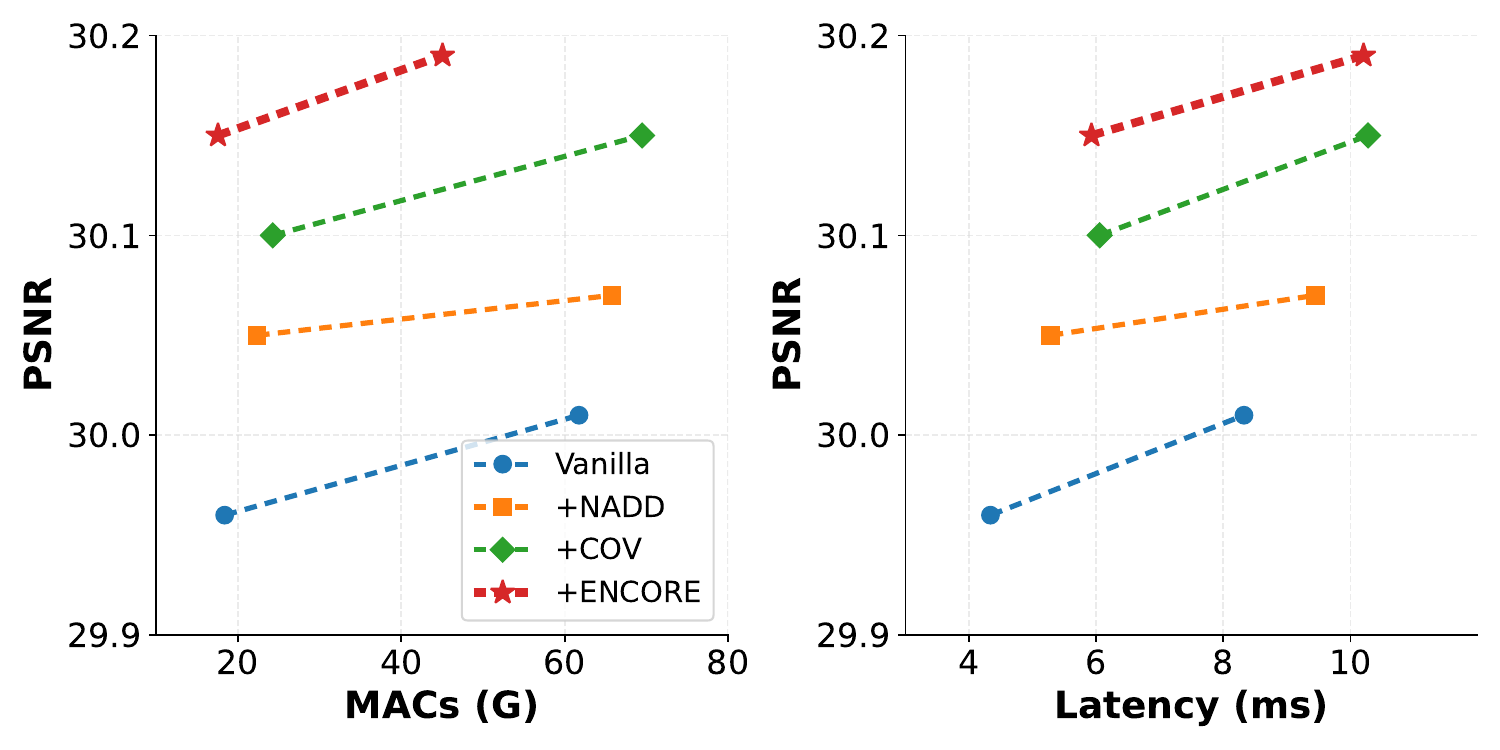}
    \caption{Performance comparison of PSNR against MACs (left) and latency (right) evaluated on Mayo2016 dataset at a 10\% dose level. 
    Each method is evaluated on two UNet variants of different model scales.}
    \label{fig:comparison_unet}
\end{figure}

As visually highlighted in \cref{fig:comparison_unet}, rather than relying on conventional parameter scaling to achieve performance gains, our ENCORE demonstrates that integrating noise context serves as a more effective paradigm for low-dose CT denoising.
Specifically, our approach allows lightweight models to achieve superior or competitive image fidelity compared to much larger baselines, while full-stack pipeline optimization substantially reduces practical latency beyond theoretical reductions in multiply-accumulate operations (MACs).
In summary, our primary contributions are listed as follows: 
\begin{itemize}
    \item We reformulate a self-supervised training pipeline based on the Cornish-Fisher expansion, providing a skewness-aware noise synthesis scheme that better matches Poisson statistics than the Gaussian approximation.
    \item We propose noise autocovariance preprocessing for model-friendly noise context estimation.
    By customizing the overall kernels, we minimize the computational overhead while boosting throughput several-fold.
    \item We propose the FlyingConv module, which adaptively adjusts the weight values based on the local anatomy and noise context information. 
    This dynamic modulation enhances both denoising quality and zero-shot robustness with minimal inference latency overhead.
\end{itemize}

\section{Related Works}
\subsection{Noise Synthesis for CT Imaging}

In X-ray CT imaging, noise in the reconstructed image is fundamentally non-stationary and spatially correlated.
While anatomy-dependent attenuation introduces non-stationary noise variance, backprojection induces complex spatial correlations, causing the noise power spectrum (NPS) to vary across local regions \cite{baek2010noise}.
To analyze these complex properties, pixel-wise noise statistics estimation from projection data has been proposed \cite{wang2017pixel}.
Because acquiring clinically clean, noise-free CT scans is infeasible, training DL-based denoising models heavily relies on synthesizing realistic training pairs.
Conventionally, supervised models are trained on synthesized LD-to-ND pairs by injecting simulated noise into clinical ND targets;
however, they tend to inherit the residual noise remaining in the ND targets, limiting their denoising performance \cite{half2half}.
To avoid inheriting this residual noise, Half2Half \cite{half2half} generates training pairs from a single CT scan while ensuring statistical independence.
In a related line of work, a training pair generation method based on a Gaussian approximation of the Poisson quantum noise has been proposed \cite{CTnoise2noise}, offering the flexibility to simulate arbitrary dose levels.

\subsection{CT Image Denoising Methods}
Most studies on DL-based CT denoising have focused on preserving fine structures and preventing over-smoothing.
To this end, several studies have designed specialized loss functions to enhance texture realism and sharpness \cite{han2021low, yang2018low, johnson2016perceptual}. 
Although modifying loss functions improves visual quality, it still poses a risk of introducing artifacts or hallucinations \cite{antun2020instabilities, eulig2024may}.
This risk arises because such loss functions primarily focus on the visual appearance or perceptual features of the output image, without explicitly modeling the underlying CT noise characteristics. 
To bridge this gap, NADD \cite{NADD} has attempted to incorporate noise power and correlation information into the denoising model.
However, NADD suffers from training instability and may not fully exploit the augmented noise properties.
To address these limitations, a pre-processing stage is required to transform noise information into a model-friendly format.
Simultaneously, the fundamental model architecture can be redesigned to adaptively reflect noise properties that are highly dependent on patient anatomy and scan geometry.
Accordingly, we propose a comprehensive denoising pipeline to maximize both denoising quality and computational efficiency.

\subsection{Modifying Convolution Operation for Denoising}
To achieve superior denoising performance and high computational efficiency, there have been various attempts to modify the fundamental convolution operation.
As pioneering approaches, bilateral filtering and guided filtering have long been widely used in natural image denoising to prevent over-smoothing around edges by adaptively adjusting the filter weights based on the local image information \cite{tomasi1998bilateral, he2012guided}.
This core concept of adaptive filtering has carried over to DL-based denoising architectures.
For instance, Malleable Convolution (MalleConv) improves computational efficiency and representation power by dynamically varying convolution weights depending on the image context \cite{malleable}.
This adaptive mechanism offers improved image quality at an equivalent computational complexity compared to conventional convolution layers that rely on static weights. 
Inspired by these prior works, we aim to maximize the denoising quality while minimizing computational overhead by dynamically varying weights based on the local characteristics.
In particular, while previous methods rely solely on local image context, our proposed FlyingConv incorporates estimated noise context information. 

\begin{figure}
    \centering
    \includegraphics[width=\textwidth]{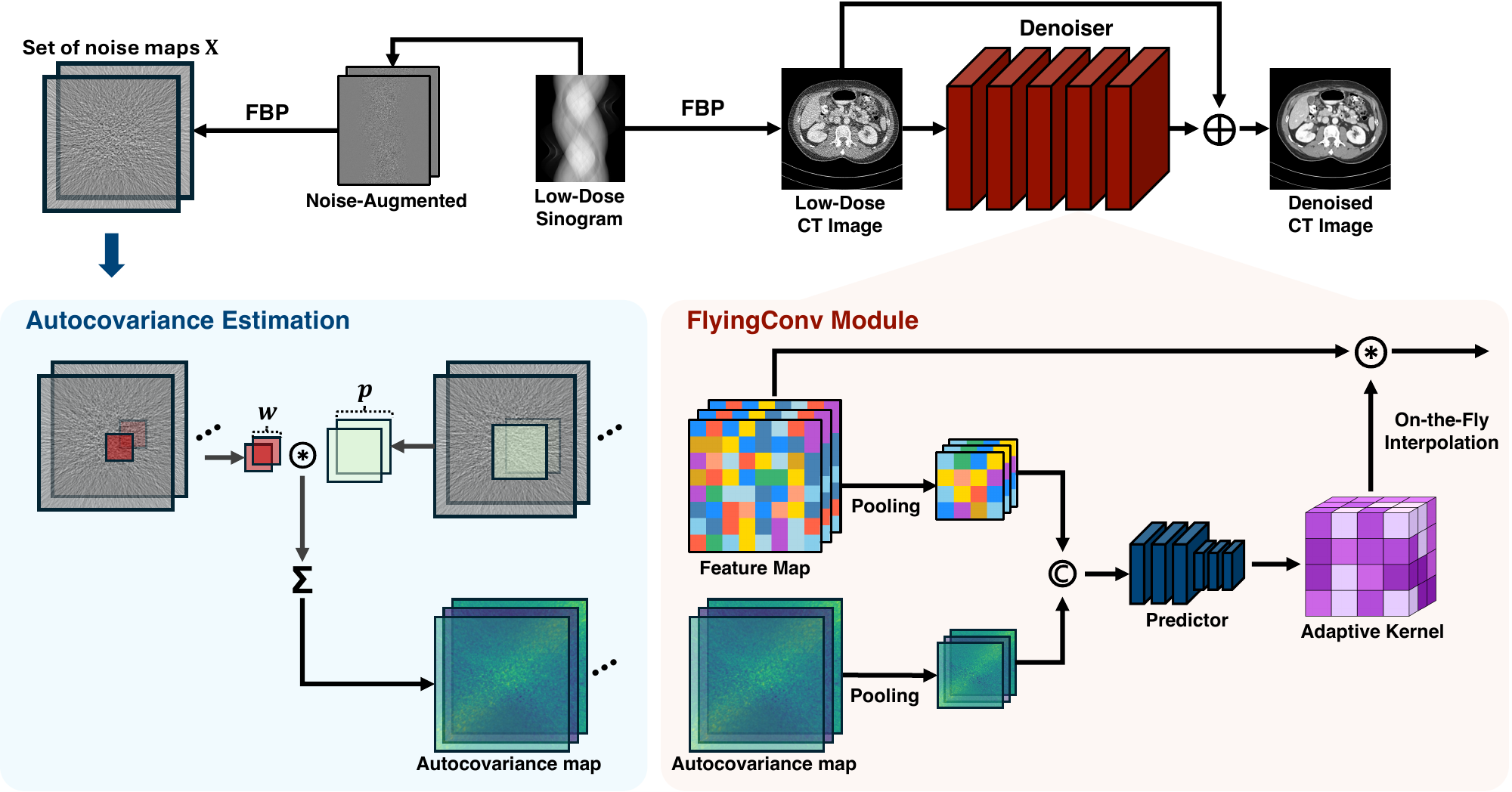}
    \caption{Overall scheme of our proposed ENCORE framework for low-dose CT denoising.
    The ENCORE framework is built upon two core components: noise autocovariance estimation and FlyingConv modules constituting the denoising model.} 
    \label{fig:main}
\end{figure}

\section{Methods}
The proposed denoising framework begins with noise augmentation using the given low-dose projection data.
This is followed by the estimation of the local noise autocovariance and its integration into the FlyingConv module.
This overall scheme is illustrated in \cref{fig:main}, where each shaded region  represents a key contribution of this work.

\subsection{Noise Synthesis for Context Integration}
In general, CT noise can be modeled as a combination of quantum noise and electronic noise, which follow Poisson and Gaussian distributions, respectively \cite{whiting2006properties}.
In ND settings, the number of detected photons $P_{\text{ND}}$ can be modeled as:
\begin{equation}
    P_{\text{ND}} = \operatorname{Pois}(P_{\text{clean}}) + \mathcal{N}(0, \sigma_\text{e}^2),
\label{eq:nd_noise_model}
\end{equation}
where $P_{\text{clean}}$ is the expected number of detected photons, $\operatorname{Pois}(P_{\text{clean}})$ represents the photon count corrupted by quantum noise, and $\mathcal{N}(0, \sigma_\text{e}^2)$ represents the Gaussian electronic noise.
Estimating $P_{\text{LD}}$ for a target dose reduction level $d$ from a given $P_{\text{ND}}$ can be approximated by adding noise to the $P_{\text{ND}}$ as follows: 
\begin{equation}
    \begin{aligned}
        P_{\text{LD}} &= \operatorname{Pois}(dP_{\text{clean}}) + \mathcal{N}(0, \sigma_\text{e}^2) \\
        &\approx d (P_{\text{ND}} + a (\operatorname{Pois}(P_{\text{ND}}) - P_{\text{ND}} + b \mathcal{N}(0, \sigma_\text{e}^2))), \\
        &\text{where } a = \sqrt{1/d - 1} \text{ and } b = \sqrt{1/d + 1}.
    \end{aligned}
\label{eq:ld_noise_model}
\end{equation}

\subsubsection{Training pair generation and noise augmentation}
To enable N2N training from a single ND scan data, a paired dataset consisting of $P_{\text{sLD}}$ and $P_{\text{ID}}$ is generated. 
Here, $P_{\text{sLD}}$ is formulated to emulate the target LD distribution, while $P_{\text{ID}}$ serves as a statistically independent counterpart to $P_{\text{sLD}}$.
By approximating the Poisson quantum noise as Gaussian noise \cite{CTnoise2noise}, this training pair can be formulated as follows:
\begin{equation}
    \begin{aligned}
        P_{\text{sLD}} &= d (P_{\text{ND}} + a (\mathcal{N}(0, P_{\text{ND}}) + b \mathcal{N}(0, \sigma_\text{e}^2))) \\
        P_{\text{ID}} &= d (P_{\text{ND}} - \frac{1}{a} (\mathcal{N}(0, P_{\text{ND}}) + \frac{1}{b} \mathcal{N}(0, \sigma_\text{e}^2))).
    \end{aligned}
\label{eq:n2n_generation}
\end{equation}

Subsequently, a set of $N$ independent noise maps $\mathbf{X}$ is constructed, where each map $\mathbf{X}_n$ estimates the difference between the low-dose image $I_{\text{LD}}$ and the ideal clean image $I_{\text{clean}}$. 
Using only the available $P_{\text{LD}}$ (or $P_{\text{sLD}}$ during training to prevent train-test discrepancy), $\mathbf{X}_n$ is formulated as follows:
\begin{equation}
\begin{aligned}
    \mathbf{X}_n &\sim \operatorname{FBP}(-\log(P_{\text{LD}} / d P_{\text{clean}})) \\ 
    &\approx \operatorname{FBP}(-\log(P_{\text{Lower}} / P_{\text{LD}})), \\
    &\text{where } P_{\text{Lower}} = P_{\text{LD}} + \mathcal{N}(0, P_{\text{LD}}) + \mathcal{N}(0, \sigma_\text{e}^2), \\
    &\text{for } n = 1, \dots, N.
\end{aligned}
\label{eq:noise_augmentation}
\end{equation}
where $P_{\text{Lower}}$ denotes the simulated lower-dose projection, and $\operatorname{FBP}$ denotes the 2D FBP reconstruction.
Note that simulated photon counts in all noise models (\cref{eq:nd_noise_model,eq:ld_noise_model,eq:n2n_generation,eq:noise_augmentation}) are clipped to $P \ge 1$ prior to logarithmic transformation, preventing non-positive photon counts as in real CT acquisitions.

However, both formulations in \cref{eq:n2n_generation,eq:noise_augmentation} hinge on the Gaussian approximation of Poisson quantum noise. 
This assumption fails in LD settings or when the scanned anatomy contains dense structures (e.g., thick bones) that severely attenuate photons and lead to photon starvation.
This approximation in turn induces a statistical discrepancy between the synthesized $P_{\text{sLD}}$ used in N2N-based training and the actual $P_{\text{LD}}$ encountered in clinical practice.
To overcome this, we apply a Cornish-Fisher expansion \cite{cornish1938moments}, a mathematical technique that transforms Gaussian variables to match the skewness of the target distribution.
To be specific, we reformulate the Gaussian noise term $\mathcal{N}(0, P)$ used in \cref{eq:n2n_generation} into a skewness-corrected term $\mathcal{W}(0, P)$.
This allows us to explicitly inject the physical skewness of the Poisson distribution into the training pair generation while still satisfying statistical independence.
The detailed derivation of $\mathcal{W}(0, P)$ is provided in the \hyperref[appendix]{Appendix}.

\subsubsection{Inference-time extension for zero-shot control}
Focusing on the model's capability to locally guide its denoising level based on the estimated noise context, we propose extending the noise augmentation formulation \cref{eq:noise_augmentation}.
During inference, this extension enables zero-shot control over the residual noise level by scaling the estimated noise intensity from $P_{\text{LD}}$ according to the target dose level we want the output image to exhibit, denoted as $d_{\text{target}}$.
To this end, by enforcing the constraint $d_{\text{target}} > d$ (where $d$ is the input low-dose level), this formulation simulates a higher-dose setting, effectively scaling the noise intensity to modulate the denoising strength.
The scaled $P_{\text{Lower}}$ can be expressed using the scaling ratio $r = d/d_{\text{target}}$ as:
\begin{equation}
    P_{\text{Lower}} = P_{\text{LD}} + \mathcal{W}(0, P_{\text{LD}}(1 - r)) + \mathcal{N}(0, \sigma_\text{e}^2(1 - r^2)).
\label{eq:scaling_noise}
\end{equation}
Note that setting $d_{\text{target}}=\infty$ yields noise maps identical to \cref{eq:noise_augmentation} (with the Gaussian approximated term $\mathcal{N}$ replaced by $\mathcal{W}$), whereas $d_{\text{target}}=1$ matches the residual noise in the output image to the normal-dose level.
Thanks to this flexibility in noise augmentation, our framework can dynamically modulate the residual noise level in a zero-shot manner, allowing for interactive texture adjustments without requiring any additional training or multiple separate models.

\subsection{Noise Autocovariance Estimation}
NADD utilizes augmented noise maps $\mathbf{X}$ as model inputs without any pre-processing.
However, because noise maps vary stochastically with each simulation, feeding raw noise realizations forces the network to implicitly extract noise statistics from fluctuating data.
Such stochastic fluctuations can constrain both the performance and robustness of the denoising model.
To address this issue, we estimate autocovariance maps $\mathbf{V} \in \mathbb{R}^{p\times p \times H \times W}$ as a preprocessing step, where $p$ denotes the spatial lag patch size. 
Assuming the noise maps have a zero mean, the autocovariance maps are calculated from $\mathbf{X}_n$ as follows:
\begin{equation}
    \begin{aligned}
        \mathbf{V}&(i, j, x, y) = \frac{1}{N w^2} \sum_{n=1}^N \sum_{(u,v) \in \Omega}\mathbf{X}_n(x+u, y+v) \\
        & \cdot \mathbf{X}_n(x+u+i-\lfloor p/2 \rfloor, y+v+j-\lfloor p/2 \rfloor), \\
        & \text{where } \Omega = \{-\lfloor w/2 \rfloor, \dots, \lfloor w/2 \rfloor\}^2,
    \end{aligned}
\label{eq:correlation_function}
\end{equation}
where $w$ denotes the spatial window size defining the local neighborhood for aggregating noise samples.
Note that the noise autocovariance is strictly defined with $w=1$ and $p = \max(H, W)$. 
However, a large $N$ is required for accurate estimation, which leads to an infeasible computational burden.
To address this limitation, we leverage the localized nature of CT noise correlation and employ a small patch size $p$ to capture the essential noise spatial correlation while substantially reducing overhead.
Furthermore, while a large $N$ requires repetitive executions of the FBP reconstruction process, we set $N=1$ and increase the spatial window size $w$.
This adaptation is based on the empirical findings that CT noise can be approximated as a stationary process in a small region of interest (ROI) on the scale of a few millimeters \cite{baek2010noise}.
By default, we set $p=5, w=5$, and $N=1$ in this work.
Finally, the autocovariance map is normalized to suppress severe value variations and compress the dynamic range, especially in low-dose settings.
In detail, we apply a signed-log normalization---computing $\operatorname{sign}(\mathbf{V}) \cdot \log(1 + |\mathbf{V}|)$---which scales down extreme peak values and stabilizes the model training. 
This overall procedure is illustrated in the blue-shaded region of \cref{fig:main}.

\subsection{On-the-Fly Weight-Modulated Convolution}
Although estimating autocovariance maps $\mathbf{V}$ provides a more model-friendly representation, we observed that simply concatenating $\mathbf{V}$ to the model input is insufficient to fully exploit the underlying noise context.
CT noise power and correlation exhibit strong spatial and anatomical dependency, varying dynamically across ROIs even under a constant X-ray tube current.
Because conventional denoising models employ static weights during inference, they struggle to adapt to such spatially and anatomically varying noise contexts.

As a solution, we develop the FlyingConv module, which dynamically adjusts the convolution weights based on both anatomical features and autocovariance maps.
To achieve high computational efficiency and a large receptive field, our module backbone initially fuses $4 \times 4$ average-pooled feature maps $\mathbf{F}$ and autocovariance maps $\mathbf{V}$.
This fused representation is then processed by a lightweight Predictor---composed of two FasterNet blocks \cite{pconv}, with an intermediate $2 \times 2$ average pooling layer---followed by a $1 \times 1$ convolution layer.
Although this structure effectively predicts spatially adaptive kernels, generating per-channel dynamic weights introduces heavy memory access costs.
To mitigate this overhead, we adopt a grouped depthwise convolution strategy that shares each predicted weight across $g$ consecutive input-output feature channels ($g=2$ by default).
Consequently, the Predictor yields a compact adaptive kernel weight tensor $\mathbf{W} \in \mathbb{R}^{k^2 \times C / g \times H/8 \times W/8}$, where $C$ denotes the number of input channels.

Next, because the pooling layers introduce a resolution mismatch between the adaptive kernel and the feature maps, the adaptive kernel should be interpolated.
However, standard implementations or naive kernel fusion (as in MalleConv \cite{malleable}) still suffer from excessive memory footprint, causing bottlenecks in both inference latency and memory usage.
To overcome this fundamental hardware bottleneck, FlyingConv is designed as a hardware-algorithm co-designed operator.
Specifically, beyond simple loop fusion, we re-engineer the dynamic convolution pipeline by delegating the on-the-fly bilinear interpolation directly to hardware-native Texture Memory Units (TMUs) (detailed in \cref{sec:GPU_optimization}).
After that, a $1 \times 1$ convolution layer is incorporated to fuse feature representations across the channel dimension, thereby compensating for the limited inter-channel interaction.
This overall procedure of the FlyingConv module is illustrated in the red shaded region of \cref{fig:main}.
Although the $1 \times 1$ convolution layer is included at the end, relying solely on the FlyingConv module still tends to under-represent inter-channel relationships.
Therefore, we sequentially alternate between the FlyingConv and standard convolution layers throughout the network.

\subsection{Implementation of Acceleration Kernels for Proposed Modules}
\label{sec:GPU_optimization}
The proposed ENCORE framework encompasses the entire process from reconstruction to post-processing, with all stages executed on the GPU for hardware acceleration.
Any localized inefficiency within this pipeline can propagate and increase the total inference latency.
To prevent this, all stages are accelerated via custom kernels using C++ and CUDA \cite{nickolls2008scalable}.

While several open-source CT image reconstruction toolboxes exist (e.g., such as LEAP \cite{LEAP}, TIGRE \cite{TIGRE}, and ASTRA \cite{ASTRA}), they may not be fully optimized for maximizing GPU utilization in parallelized FBP workflows.
As a solution, we develop custom reconstruction kernels designed to minimize the computational overhead between FBP stages.
Our custom kernels are built upon the LEAP library, re-engineered to maximize GPU cache efficiency. 
They minimize the memory transfer overhead between the FBP and subsequent denoising stages. 

Similarly, the computation of noise autocovariance and the FlyingConv modules are also accelerated via custom kernels.
In our implementation, we leverage the TMUs during the on-the-fly interpolation process of FlyingConv.
Because TMUs provide hardware-accelerated interpolation, the need to explicitly access four neighboring memory addresses for a manual weighted-sum calculation is circumvented.
This hardware-level acceleration reduces memory access latency and mitigates the computational bottleneck during adaptive kernel generation.

\section{Experiments}
\subsection{Datasets}
First, we utilized the ``2016 NIH-AAPM-Mayo Clinic Low-Dose CT Grand Challenge" (Mayo2016) dataset for training and testing \cite{Mayo2016}.
We used 1-mm thickness normal-dose CT images (NDCT) from 10 patients, partitioning such that 4,773 slices from 8 patients were used as training and validation, while the remaining 1,093 slices from 2 patients served as the test dataset.
These clinical NDCT images were treated as noiseless reference images.
To simulate the raw CT data acquisition process under a monoenergetic X-ray beam assumption, we first performed fan-beam geometry forward projection on the reference images to generate ground-truth (GT) clean data $P_{\text{clean}}$.
Subsequently, simulated ND projection data $P_{\text{ND}}$ were generated by adding noise according to \cref{eq:nd_noise_model}.
Following a widely used setup in CT noise simulation studies \cite{ma2012variance}, the number of incident photons $N_{\text{in}}$ and the electronic noise variance $\sigma_\text{e}^2$ were set to $5 \times 10^5$ and $4$, respectively.
For low-dose CT image (LDCT) data, simulated LD projection data $P_{\text{LD}}$ were generated with two dose levels, $d \in \{0.25, 0.1\}$, according to \cref{eq:ld_noise_model}.
Notably, only the $d=0.25$ configuration was used for training the denoising model, leaving the $d=0.1$ configuration to evaluate the robustness of denoisers against an unseen noise distribution.

To evaluate generalization performance, we also utilized the "Low Dose CT Image and Projection Data" dataset released by the Mayo Clinic in 2020 (Mayo2020) \cite{Mayo2020}.
While the Mayo2016 dataset consists exclusively of data acquired from Siemens Healthineers scanners (SOMATOM Definition AS+ and SOMATOM Definition Flash), the Mayo2020 dataset encompasses patients scanned across multiple CT systems.
To evaluate denoising performance under cross-vendor and multi-organ conditions, we selected a total of 850 slices from 10 patients scanned with a GE Healthcare Discovery CT750 HD system.
For both Mayo2016 and Mayo2020 datasets, projections were acquired using 1024 projection views, while the remaining geometry parameters (e.g., source-to-isocenter and source-to-detector distance) were configured according to their respective DICOM header metadata.
Consequently, patient-specific variations in reconstruction pixel spacing are preserved, causing the manifested noise correlation and texture to vary across subjects even on identical scanners.

To further validate the practical utility of our method beyond simulated datasets, we also evaluated our method on real-world raw data.
We scanned an anthropomorphic chest phantom using our in-house tabletop cone-beam CT system, which is equipped with a Varex G-1593BI rotating anode X-ray tube and a Varex PaxScan 4030CB flat-panel detector.
To mitigate scatter artifacts, the X-ray beam was collimated to \qty{4}{\cm} along the axial direction during acquisition.
A total of 720 projection views were acquired in a $2 \times 2$ binning mode with a binned resolution of 1024 detector columns.
The scan was acquired at a tube voltage of \qty{120}{kVp} and a tube current of \qty{10}{\mA}, utilizing a pulsed exposure of \qty{20}{\ms} per projection.
In addition, to accurately model electronic and quantum noise characteristics, dark field, flood field, and \qty{20}{\cm} acrylic slab scans were pre-acquired for system calibration and noise augmentation.

\subsection{Training and Evaluation Details}
Our ENCORE was integrated into non-hierarchical DnCNN \cite{dncnn} and hierarchical UNet \cite{unet} backbones to validate its effectiveness compared to other variants.
In particular, we compared four configurations for each backbone: (1) the Vanilla baseline, (2) NADD integration (+NADD), (3) NADD with the noise maps replaced by the autocovariance map (+COV), and (4) ENCORE integration (+ENCORE).
When implementing ENCORE, the odd-indexed layers of the Vanilla model were replaced with the FlyingConv modules, retaining their original specifications (e.g., kernel size).
In addition, RED-CNN \cite{chen2017low}, NAFNet \cite{nafnet} and Uformer \cite{uformer} were evaluated in their Vanilla configurations to serve as representative attention-based baselines.
All models were trained for 300 epochs on $256 \times 256$ cropped patches with a batch size of 32, using the Adam optimizer \cite{kingma2015adam} with an initial learning rate of $2 \times 10^{-4}$ that was halved every 60 epochs. 
Due to the constraints of the N2N framework, the loss function was restricted to mean squared error (MSE).
To benchmark recently proposed approaches specialized for CT denoising, we evaluated CoreDiff \cite{corediff} and FoundDiff \cite{founddiff}.
While all other methods strictly operate on a single slice within the self-supervised N2N regime, both models were trained under noise-to-clean (N2C) supervision.
Because the original CoreDiff holds a substantial informational advantage by incorporating adjacent slices unlike other single-slice denoisers, we trained and evaluated two separate variants: a single-slice model and a concatenated triplet-slice model.
For FoundDiff, to preserve its claimed capability to understand anatomical context and dose levels, we initialized the model with official pre-trained weights, froze the dose- and anatomy-aware contrastive language-image pre-training (DA-CLIP) encoder, and fine-tuned only the diffusion backbone.

Except for RED-CNN and the diffusion-based models, each architecture was analyzed across two scale variants: Lite and Base configurations.
In detail, UNet variants were scaled by adjusting the number of feature channels, whereas DnCNN variants were scaled by modifying the number of intermediate layers.
We measured the MACs and actual inference latency for all configurations using an NVIDIA RTX A6000 GPU.
Note that during the latency measurement, evaluating NAFNet and Uformer using their official implementations yielded a substantially higher latency compared to the other models, even under comparable computational budgets.
This discrepancy arises because the attention mechanisms underlying both models introduce a severe memory bandwidth bottleneck.
To address memory bandwidth overheads, we applied kernel fusion via \texttt{torch.compile()} \cite{ansel2024pytorch} to all models except DnCNN and UNet prior to latency measurement.

For quantitative evaluation, we computed peak signal-to-noise ratio (PSNR), structural similarity (SSIM) \cite{ssim}, and the area under the hallucination operating characteristic curve (AUHOC) \cite{kc2026sfrc} after clipping the image values to $[-300, 300]$ Hounsfield Units (HU).
While PSNR and SSIM measure overall fidelity and structural similarity, AUHOC mainly focuses on the presence of structural hallucinations by computing patch-wise similarity in the frequency domain.
Unlike PSNR and SSIM, a lower AUHOC value signifies superior denoising performance with fewer structural hallucinations.
Regarding statistical evaluation, simply comparing $\text{mean} \pm \text{std}$ is insufficient because quality metrics vary considerably across slice locations even within the same patient. 
Therefore, to verify whether the best-performing configuration significantly outperformed the remaining variants, we conducted a paired t-Test with a patient-specific AR(1) model-based correction and Holm's $p$-value adjustment ($p < 0.05$) for multiple comparisons \cite{holm1979simple}.

\begin{figure}
    \centering
    \captionsetup[subfloat]{labelformat=empty, font=scriptsize}
    \subfloat[Noisy (LDCT)]{\includegraphics[width=0.24\textwidth]{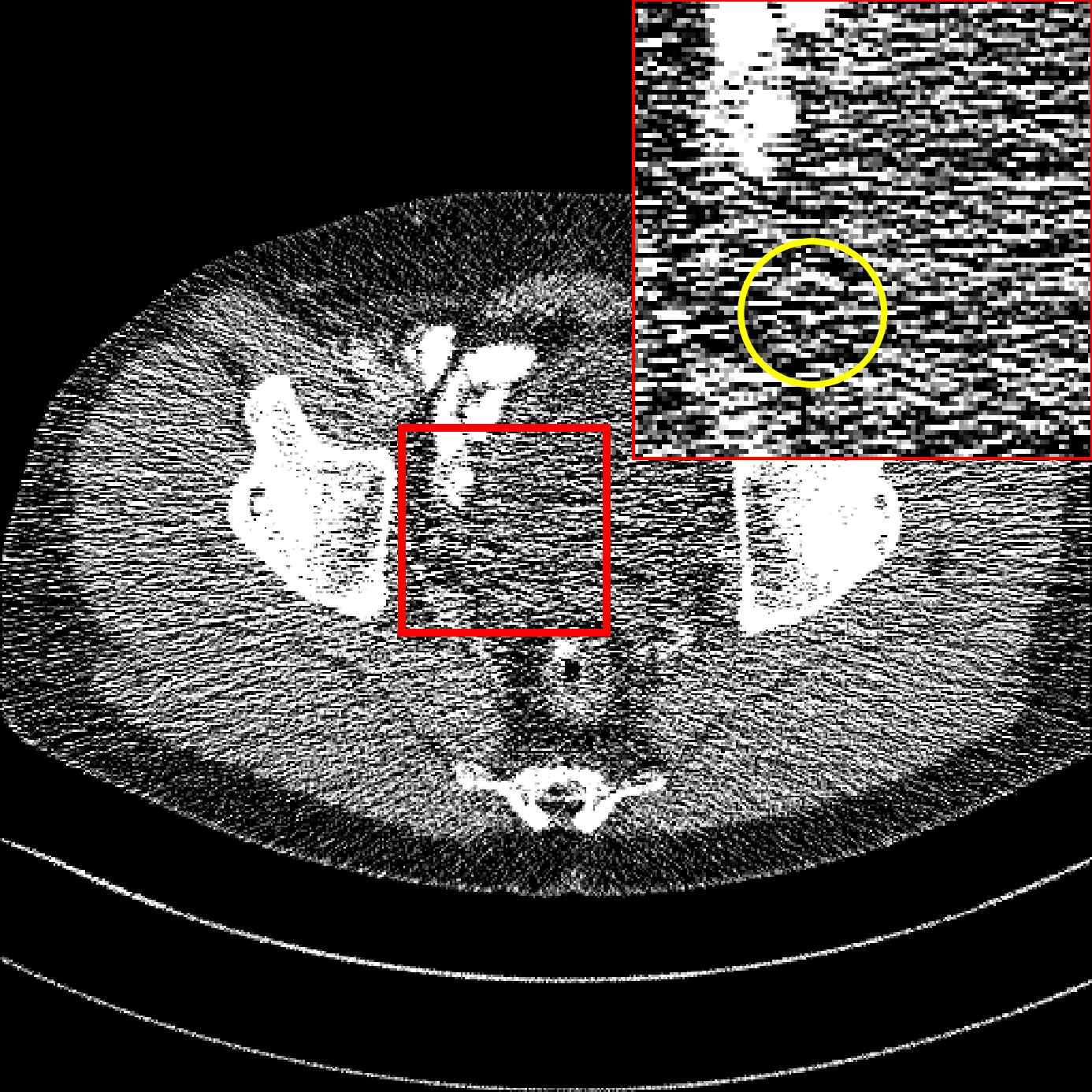}}%
    \subfloat[CoreDiff-Triplet]{\includegraphics[width=0.24\textwidth]{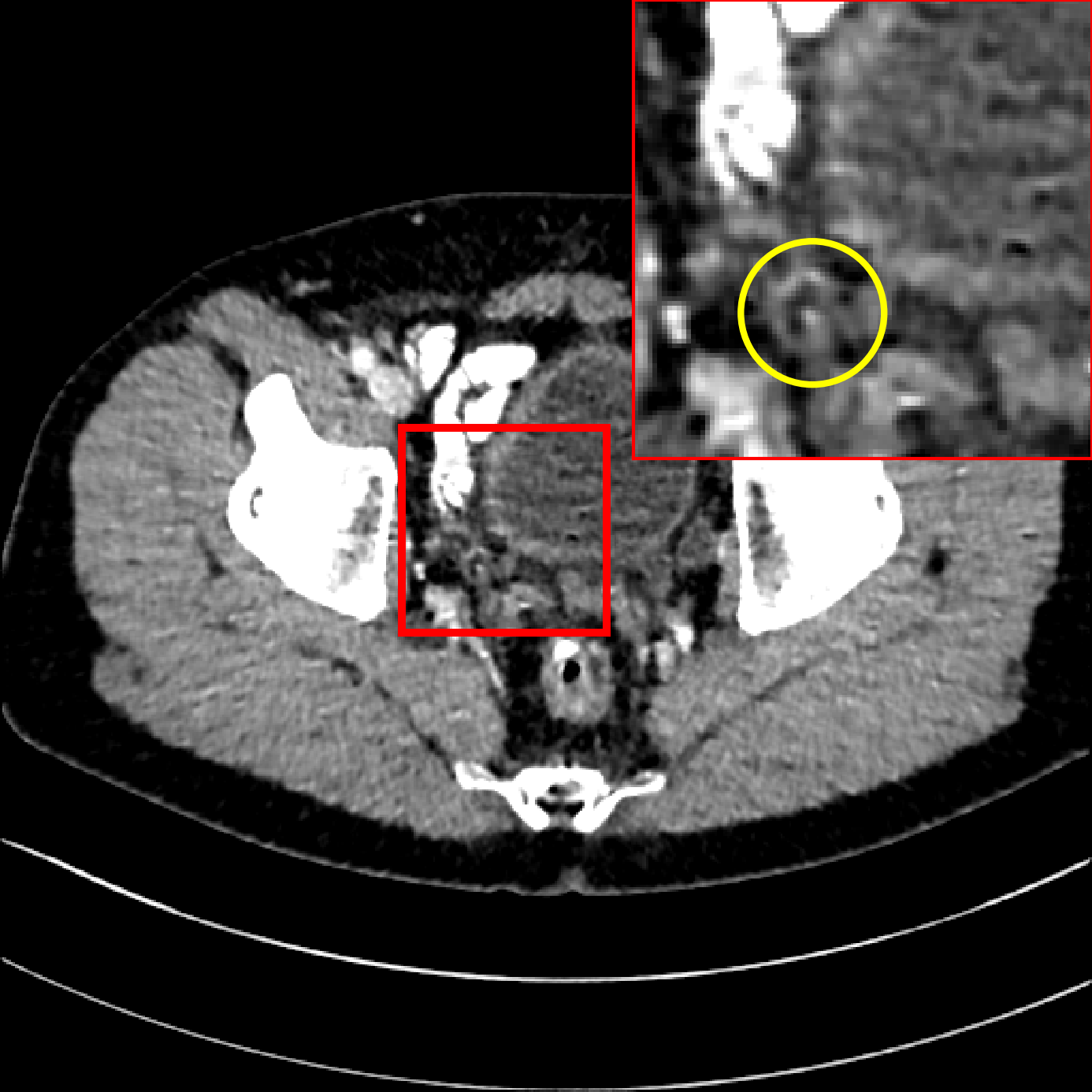}}%
    \subfloat[FoundDiff]{\includegraphics[width=0.24\textwidth]{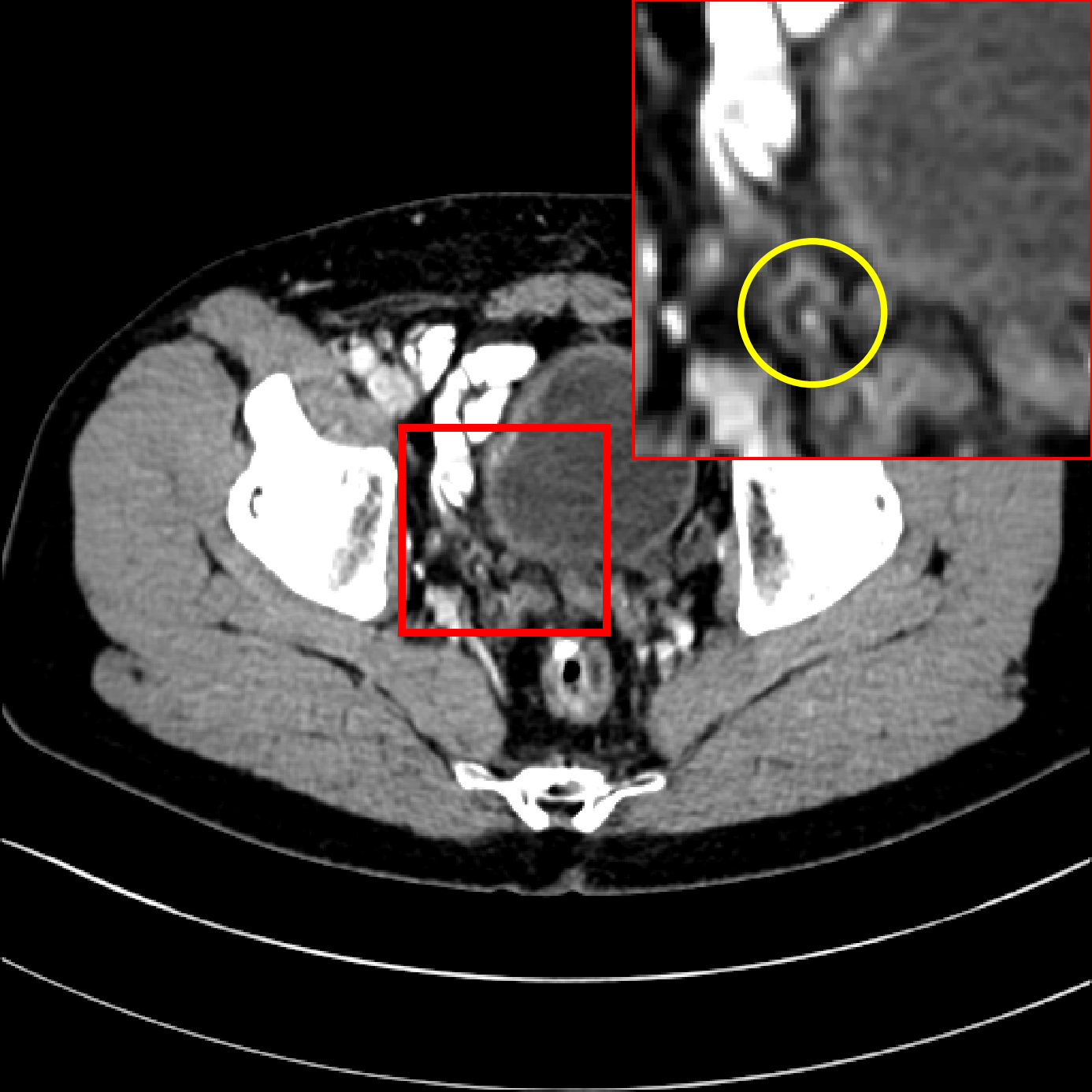}}%
    \subfloat[Ground Truth (NDCT)]{\includegraphics[width=0.24\textwidth]{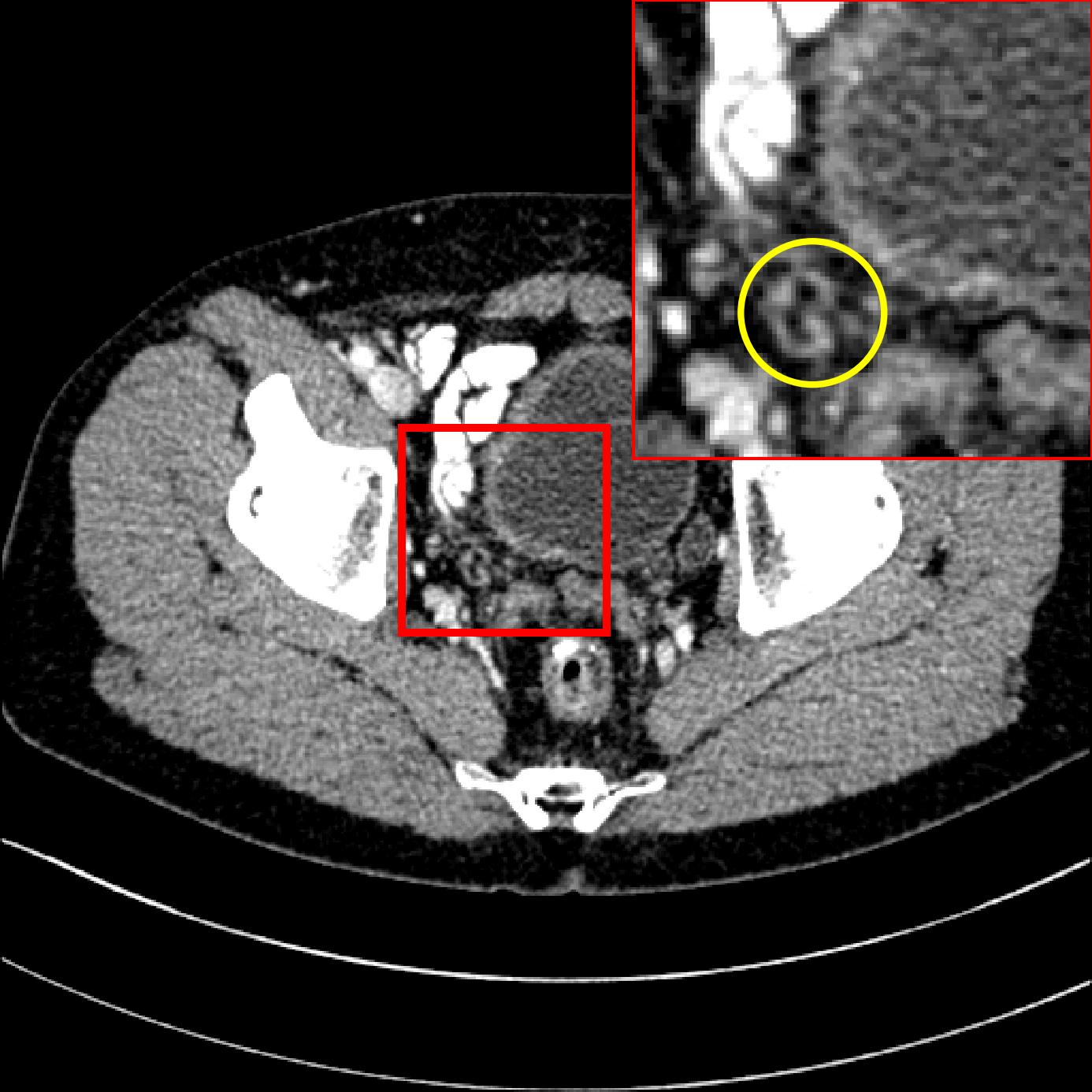}}%
    \\[-2.0ex]
    \subfloat[UNet (Vanilla)]{\includegraphics[width=0.24\textwidth]{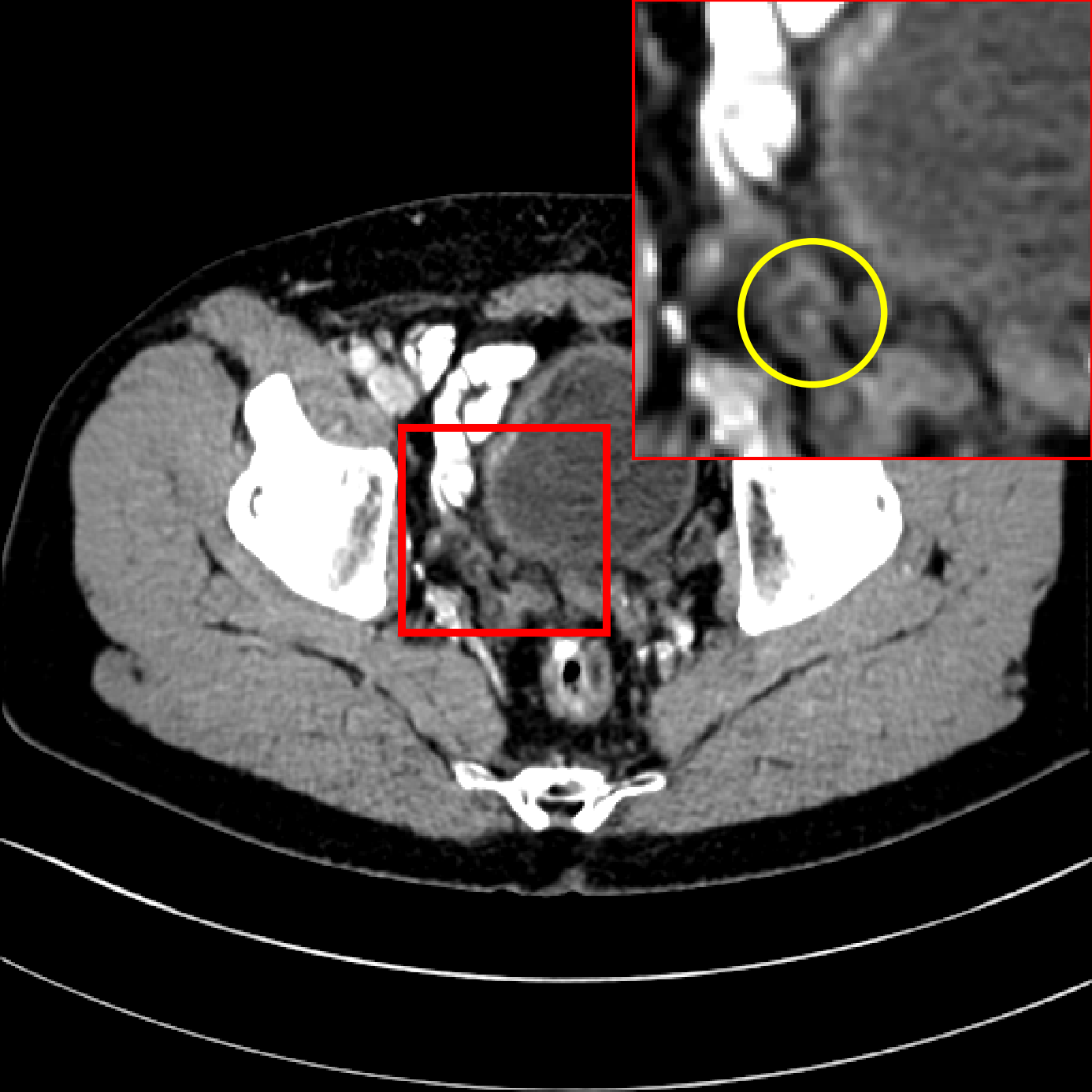}}%
    \subfloat[UNet+NADD]{\includegraphics[width=0.24\textwidth]{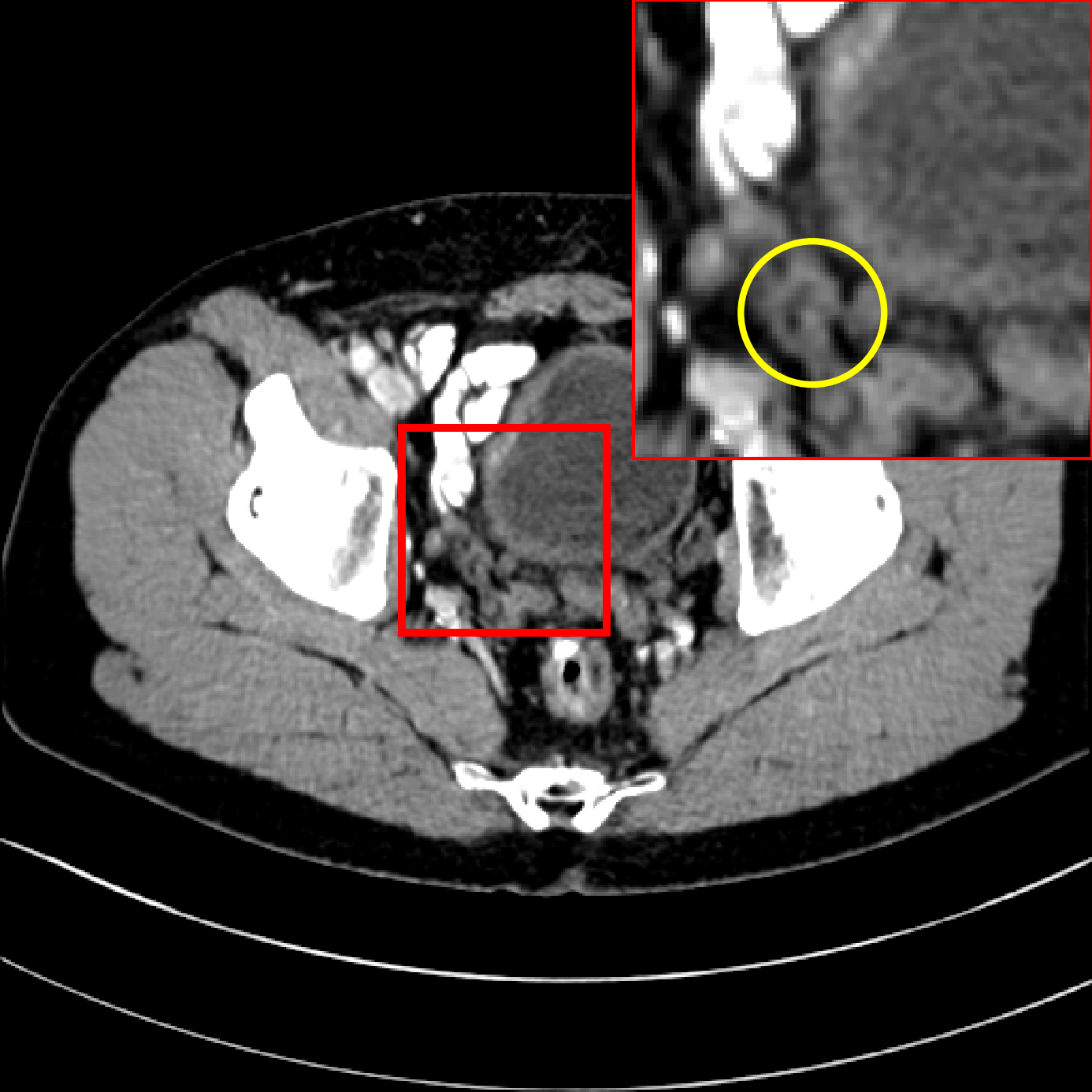}}%
    \subfloat[UNet+COV (Ours)]{\includegraphics[width=0.24\textwidth]{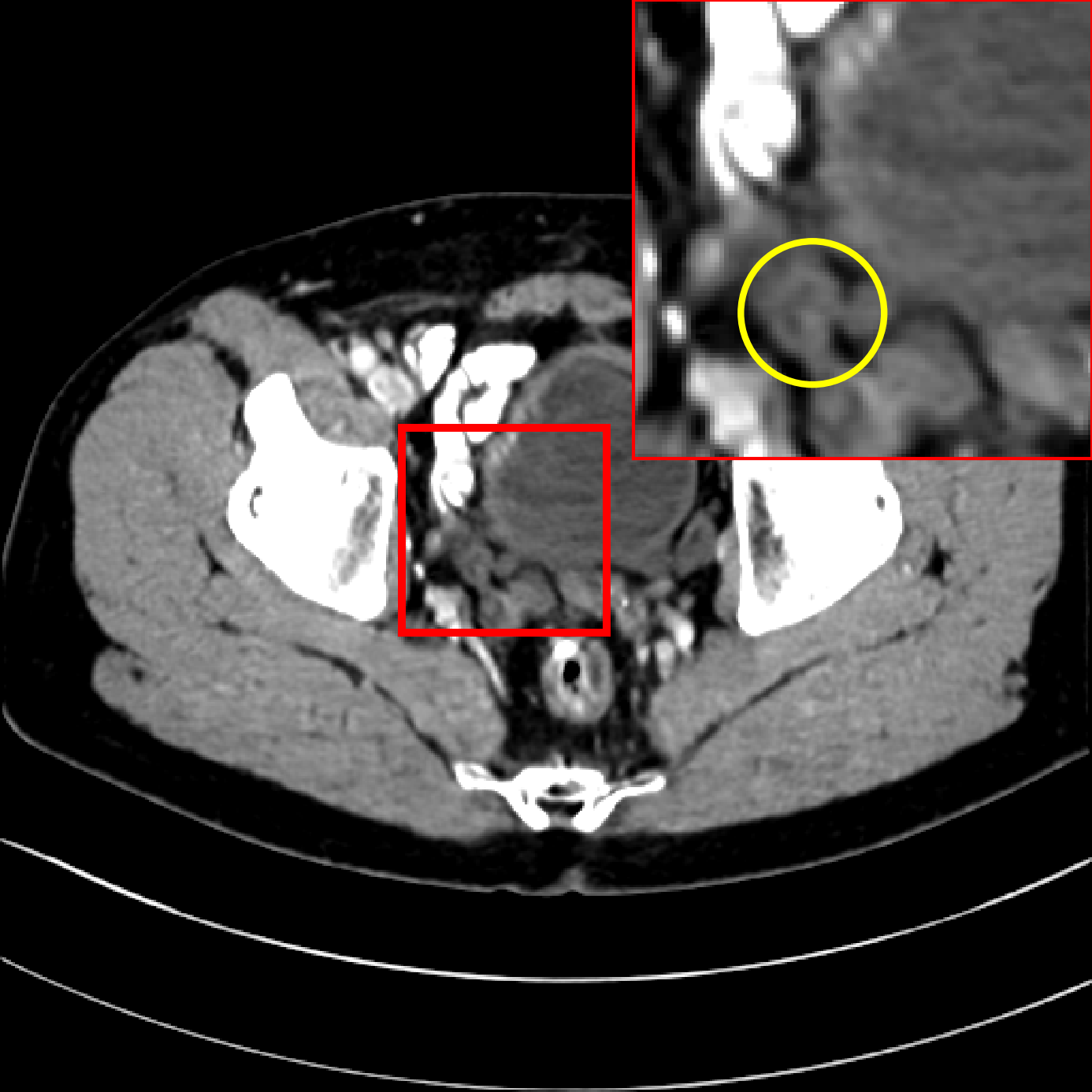}}%
    \subfloat[\textbf{UNet+ENCORE (Ours)}]{\includegraphics[width=0.24\textwidth]{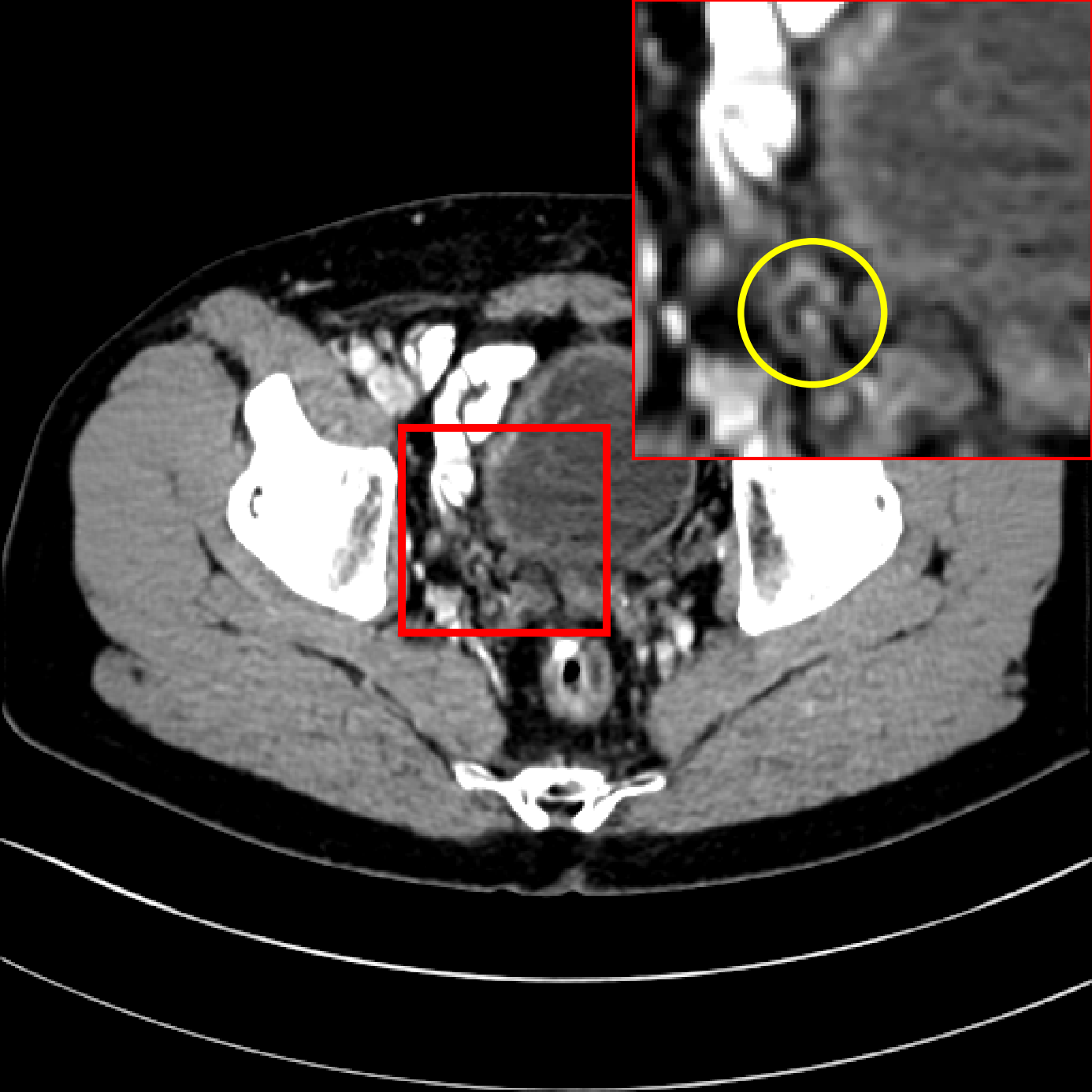}}%
    \caption{Qualitative comparison on the Mayo2016 dataset at a 10\% dose level with zoomed-in ROI patches (display window: [-100, 200]~HU). UNet variants are evaluated using the Base configuration.}
    \label{fig:qualitative_mayo2016}
\end{figure}

\begin{figure}
    \centering
    \captionsetup[subfloat]{labelformat=empty, font=scriptsize}
    \subfloat[Noisy (LDCT)]{\includegraphics[width=0.24\textwidth]{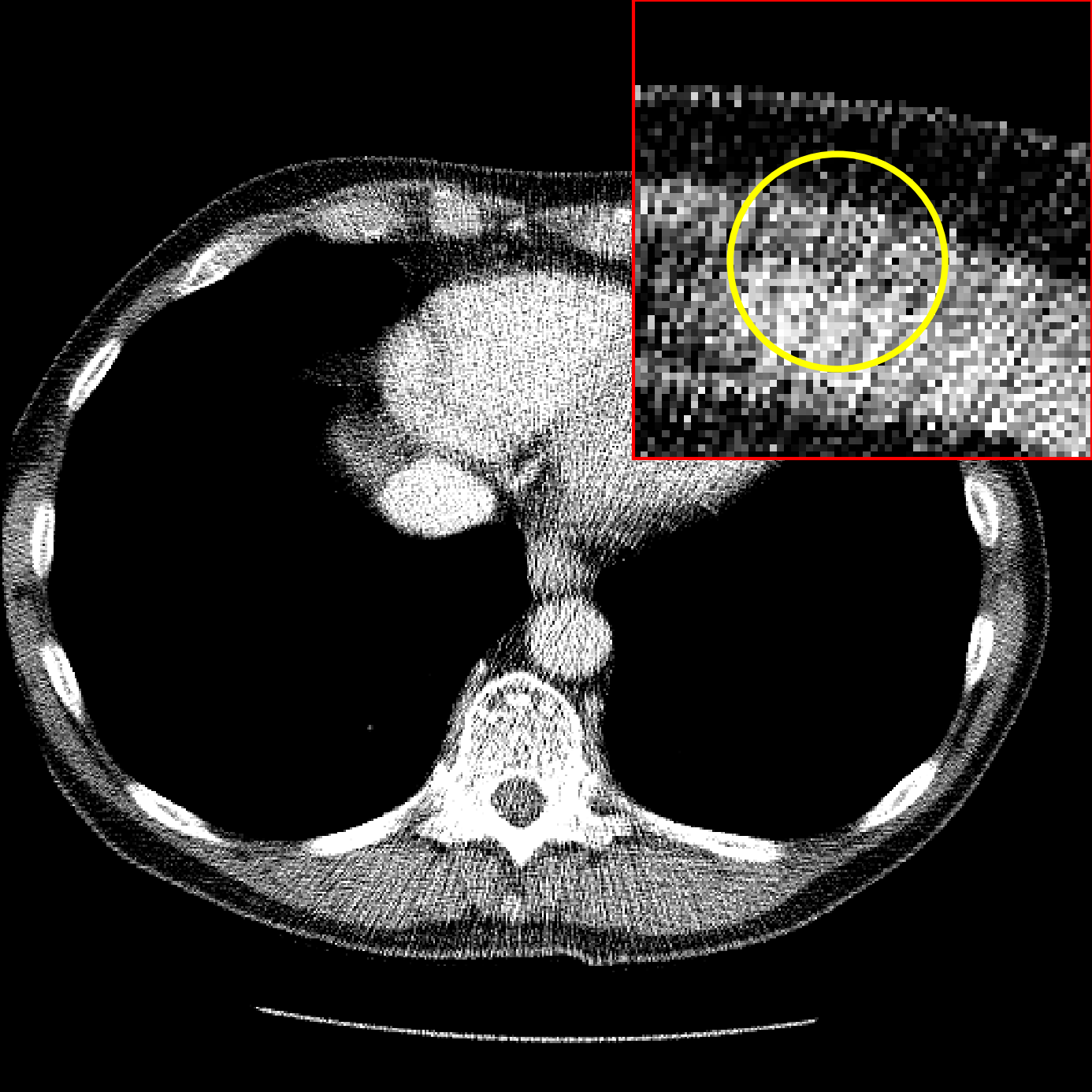}}%
    \subfloat[CoreDiff-Triplet]{\includegraphics[width=0.24\textwidth]{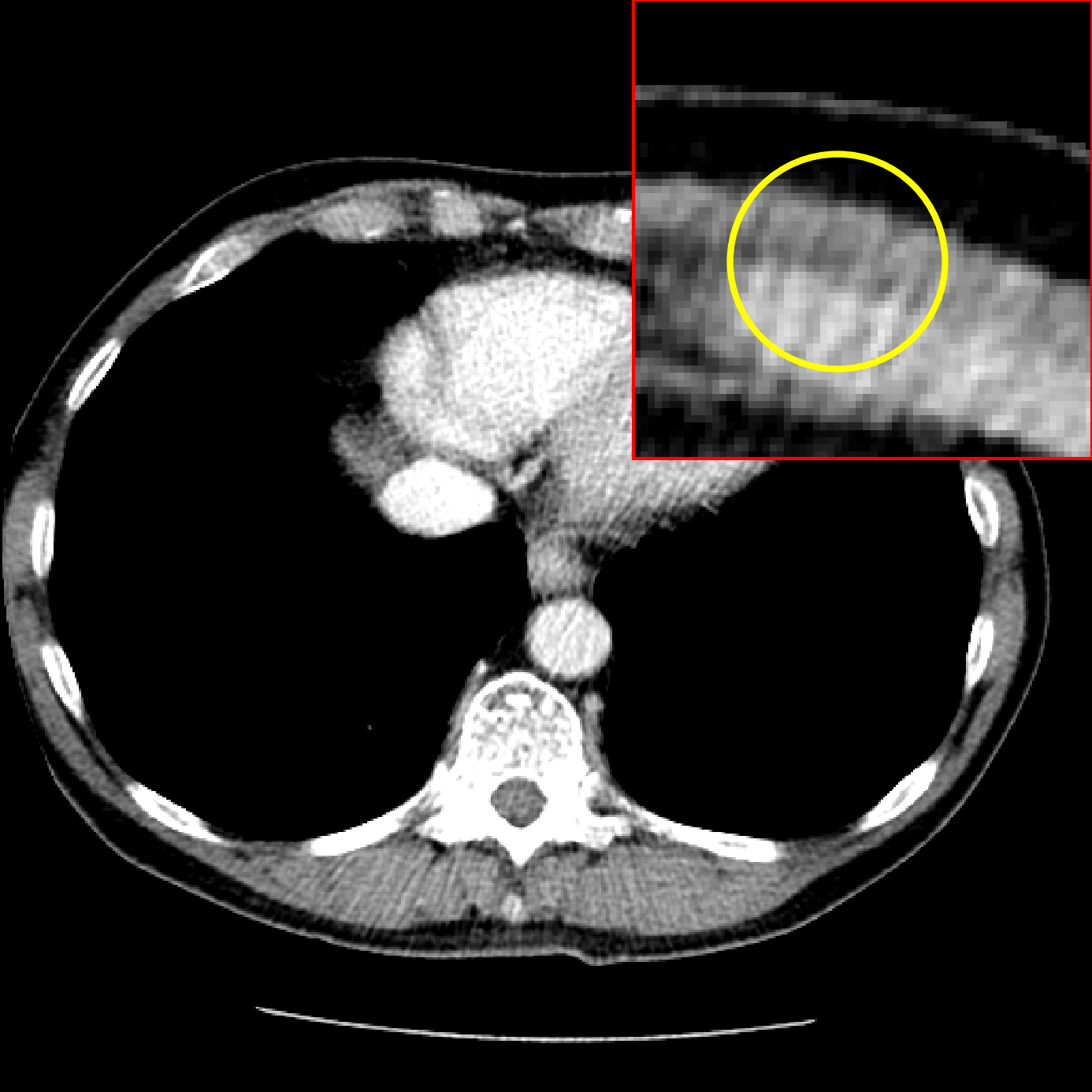}}%
    \subfloat[FoundDiff]{\includegraphics[width=0.24\textwidth]{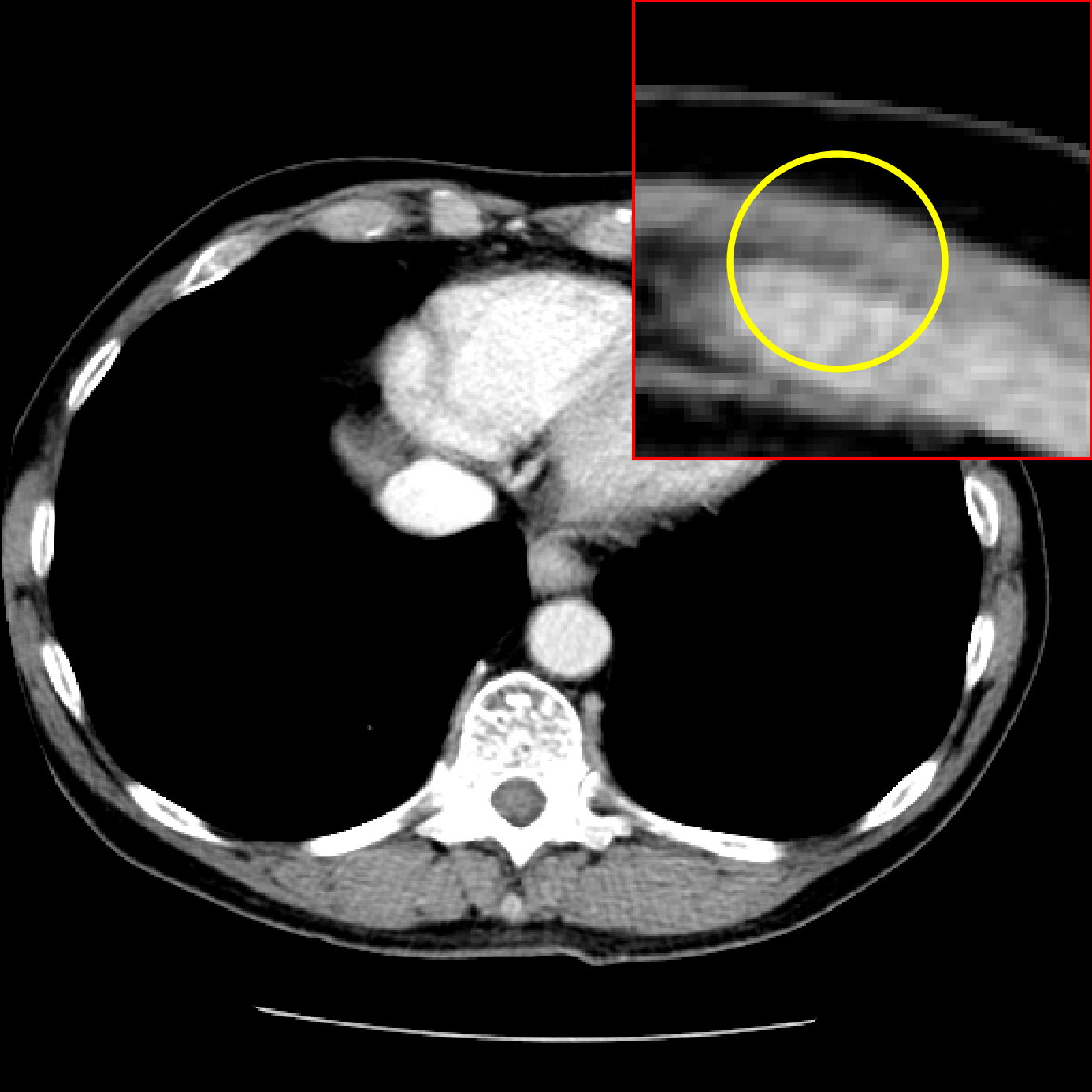}}%
    \subfloat[Ground Truth (NDCT)]{\includegraphics[width=0.24\textwidth]{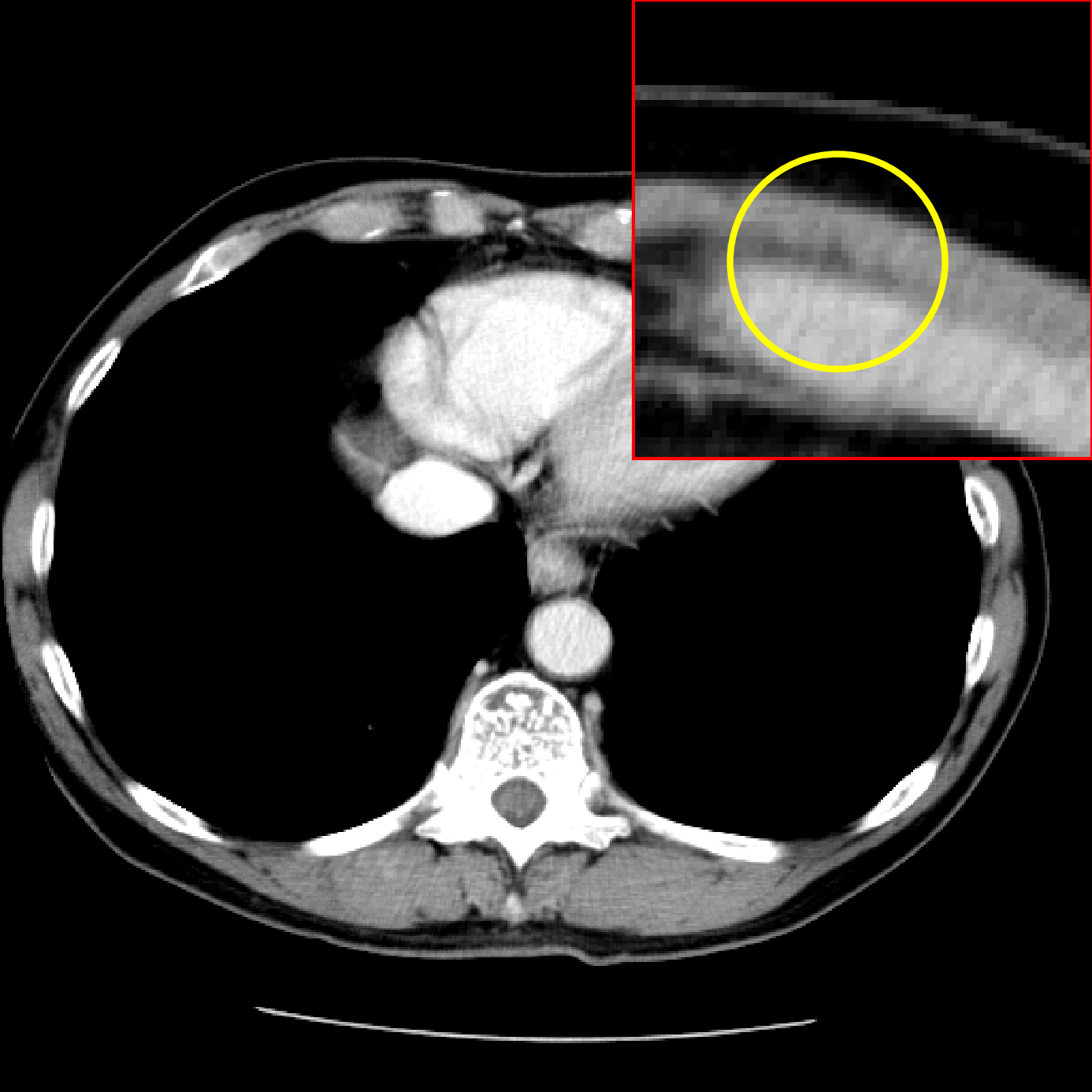}}%
    \\[-2.0ex]
    \subfloat[UNet (Vanilla)]{\includegraphics[width=0.24\textwidth]{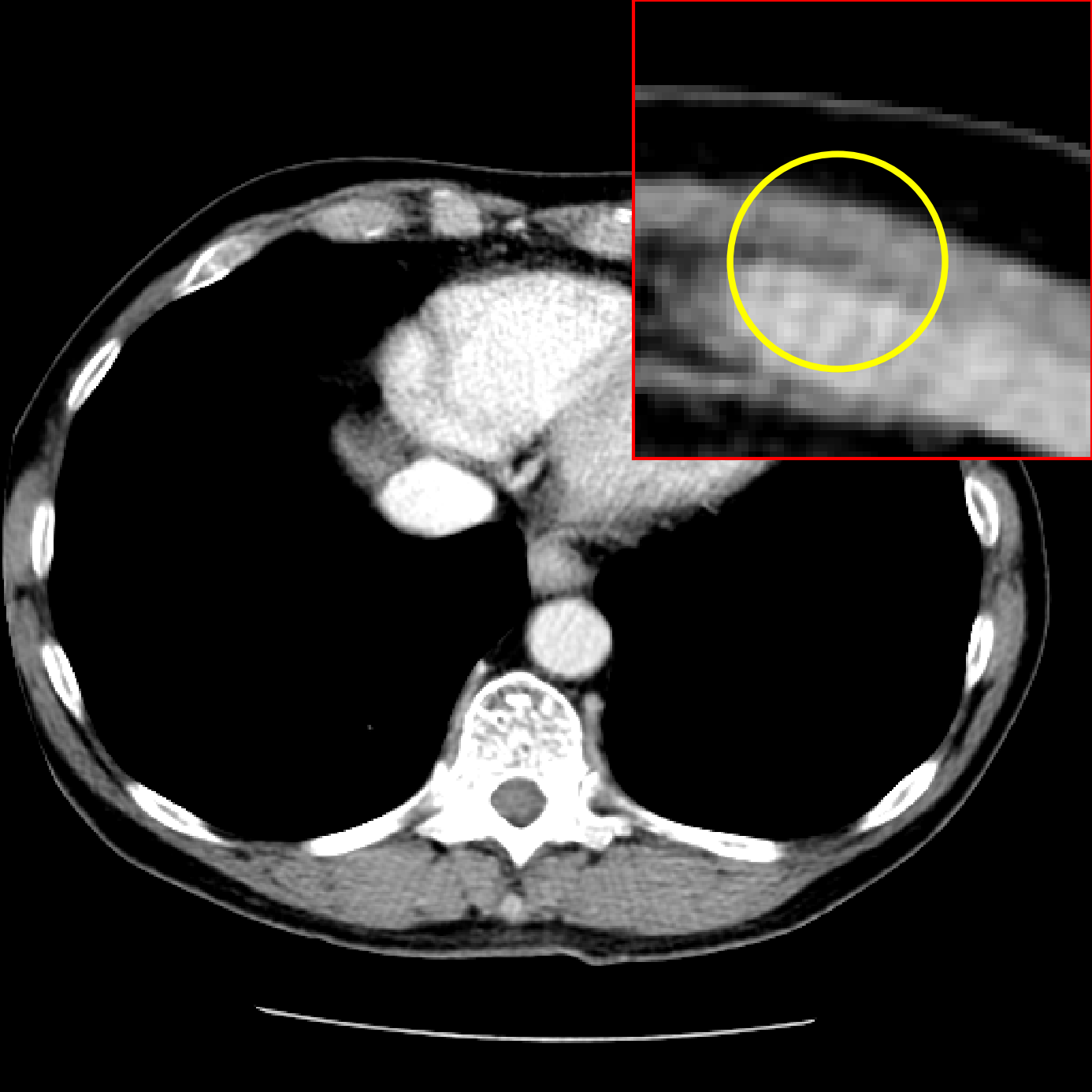}}%
    \subfloat[UNet+NADD]{\includegraphics[width=0.24\textwidth]{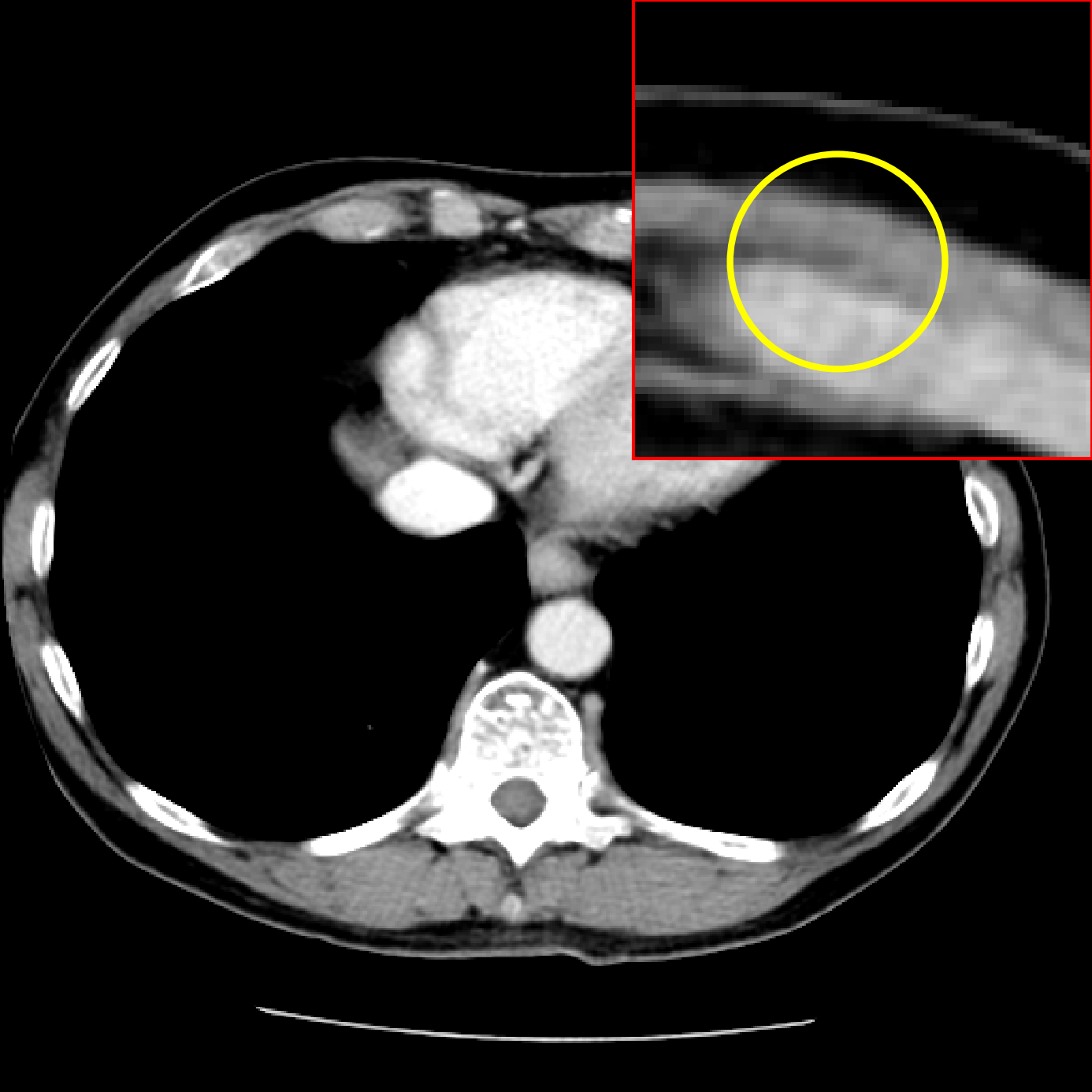}}%
    \subfloat[UNet+COV (Ours)]{\includegraphics[width=0.24\textwidth]{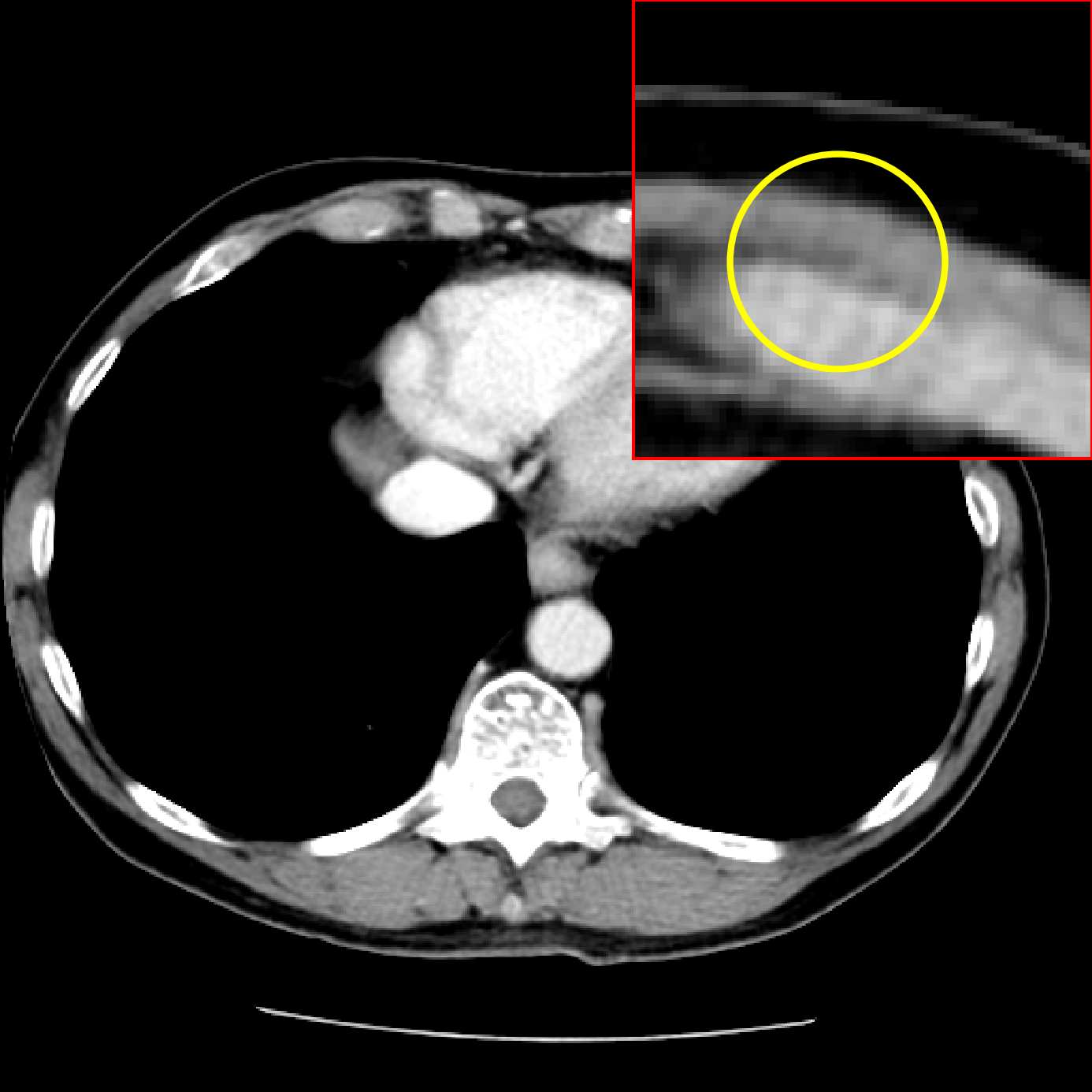}}%
    \subfloat[\textbf{UNet+ENCORE (Ours)}]{\includegraphics[width=0.24\textwidth]{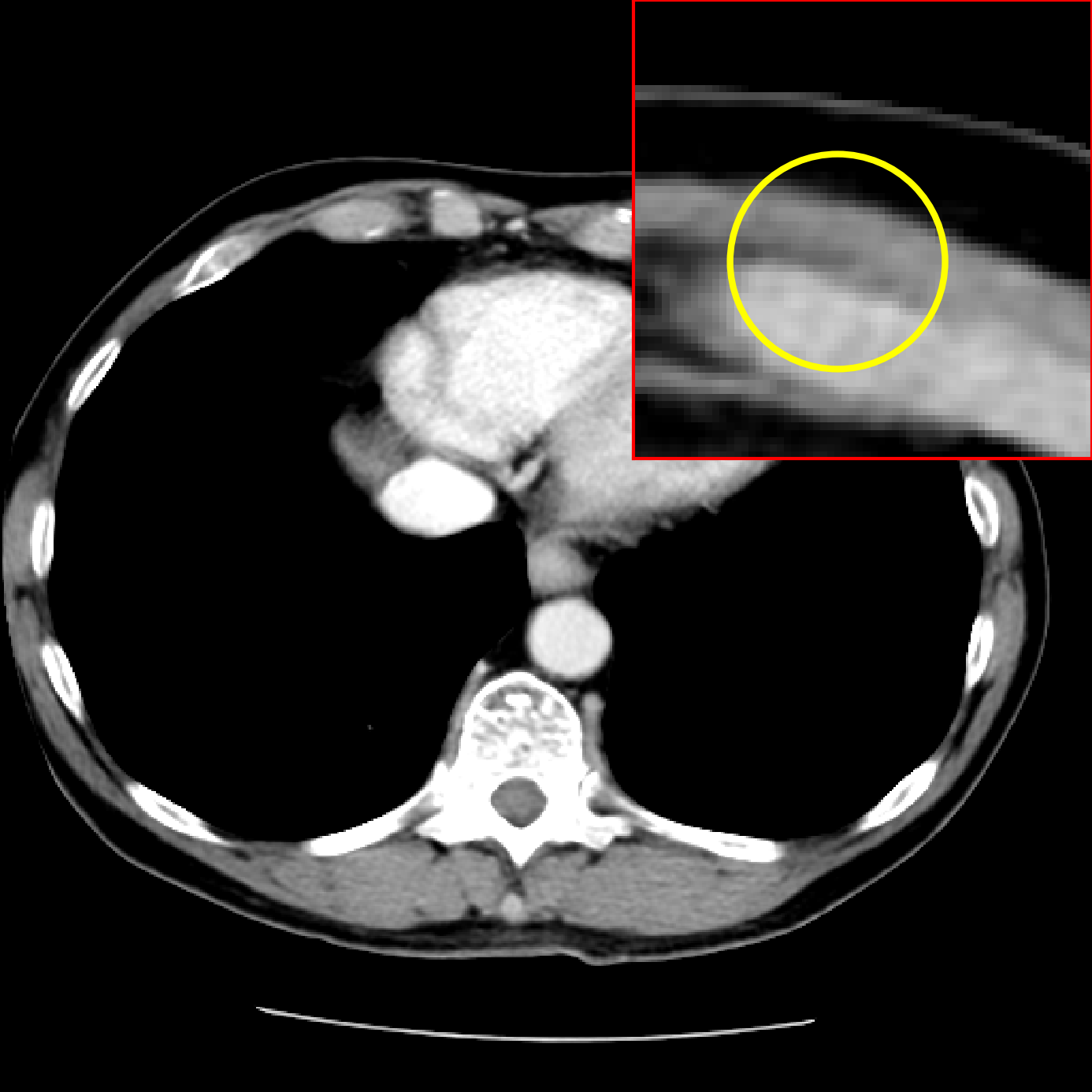}}%
    \caption{Qualitative comparison on the Mayo2020 dataset at a 10\% dose level with zoomed-in ROI patches (display window: [-100, 200]~HU). UNet variants are evaluated using the Lite configuration.}
    \label{fig:qualitative_mayo2020}
\end{figure}

\begin{figure}
    \centering
    \captionsetup[subfloat]{labelformat=empty, font=scriptsize}
    \subfloat[Noisy]{\includegraphics[width=0.24\textwidth]{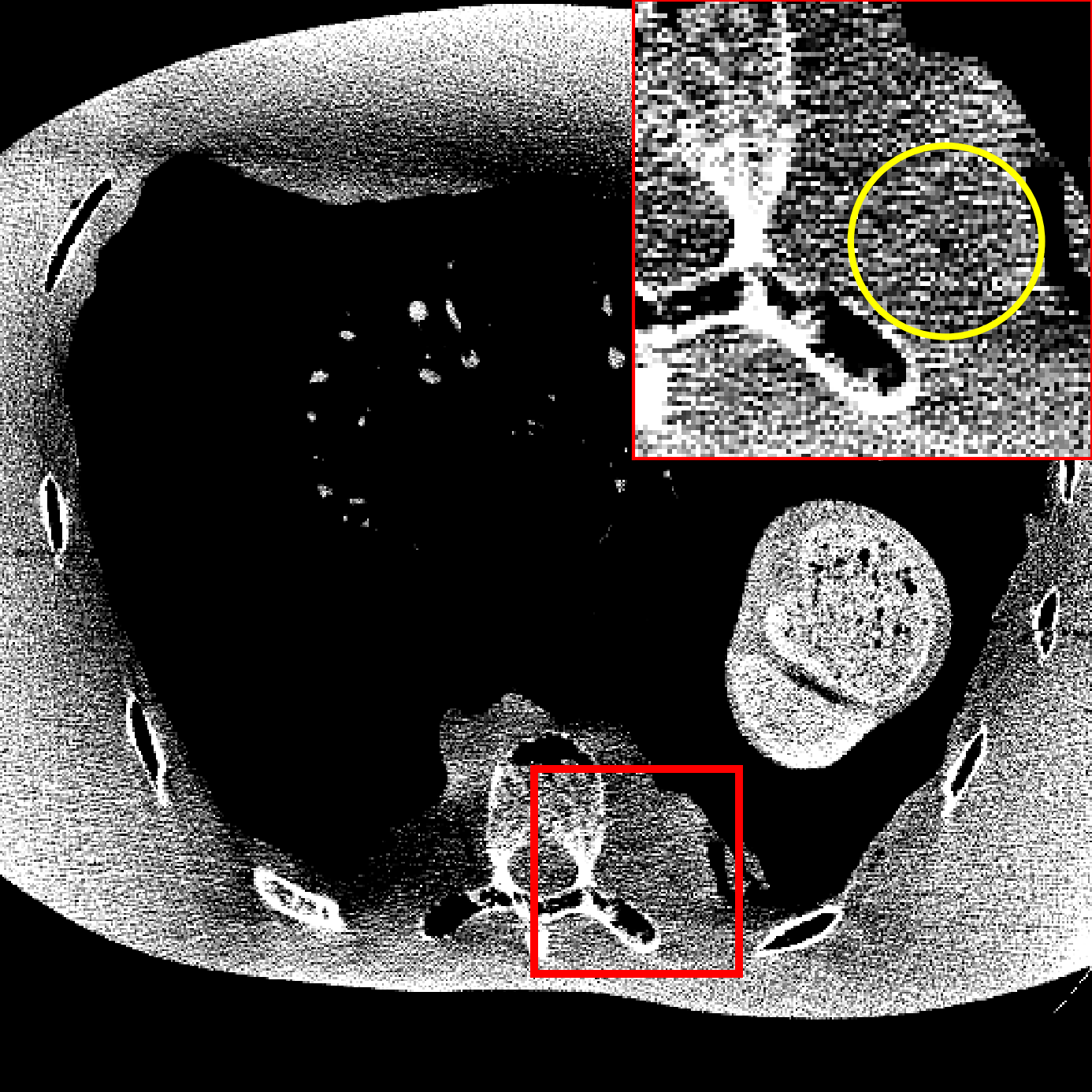}}%
    \subfloat[CoreDiff-Triplet]{\includegraphics[width=0.24\textwidth]{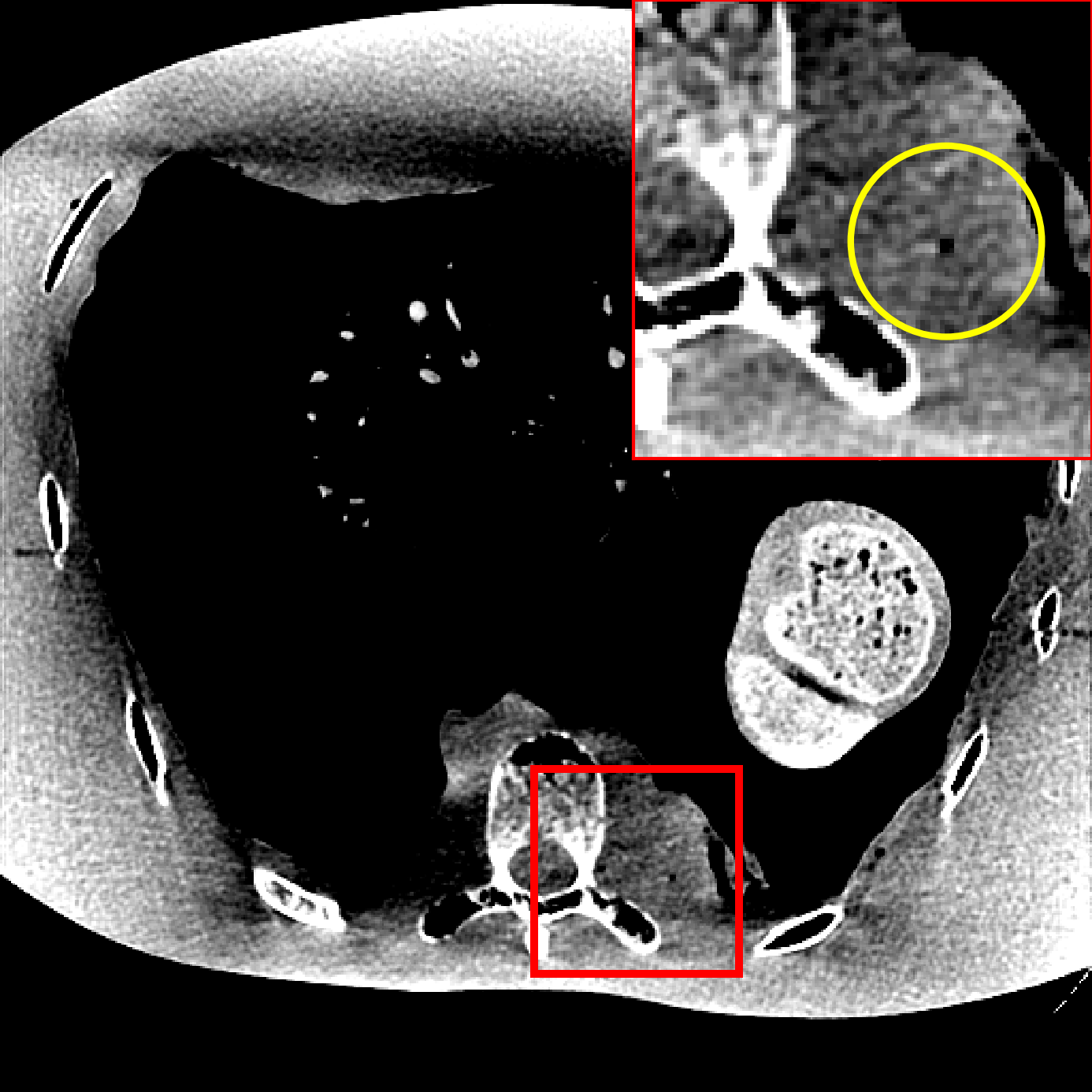}}%
    \subfloat[FoundDiff]{\includegraphics[width=0.24\textwidth]{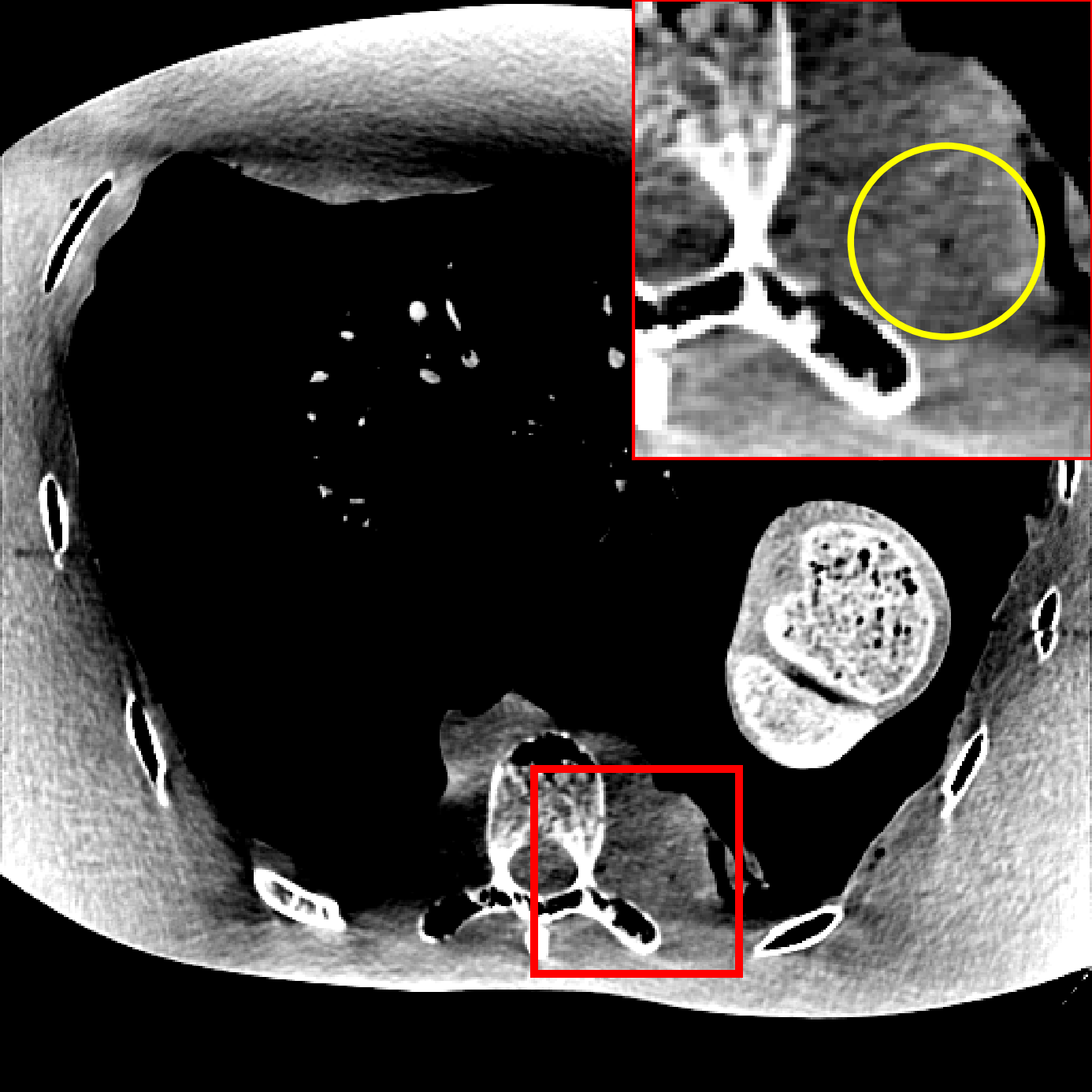}}%
    \subfloat[GT]{\makebox[0.24\textwidth]{\raisebox{0.12\textwidth}[0pt][0pt]{\Large --}}}%
    \\[-2.0ex]
    \subfloat[UNet (Vanilla)]{\includegraphics[width=0.24\textwidth]{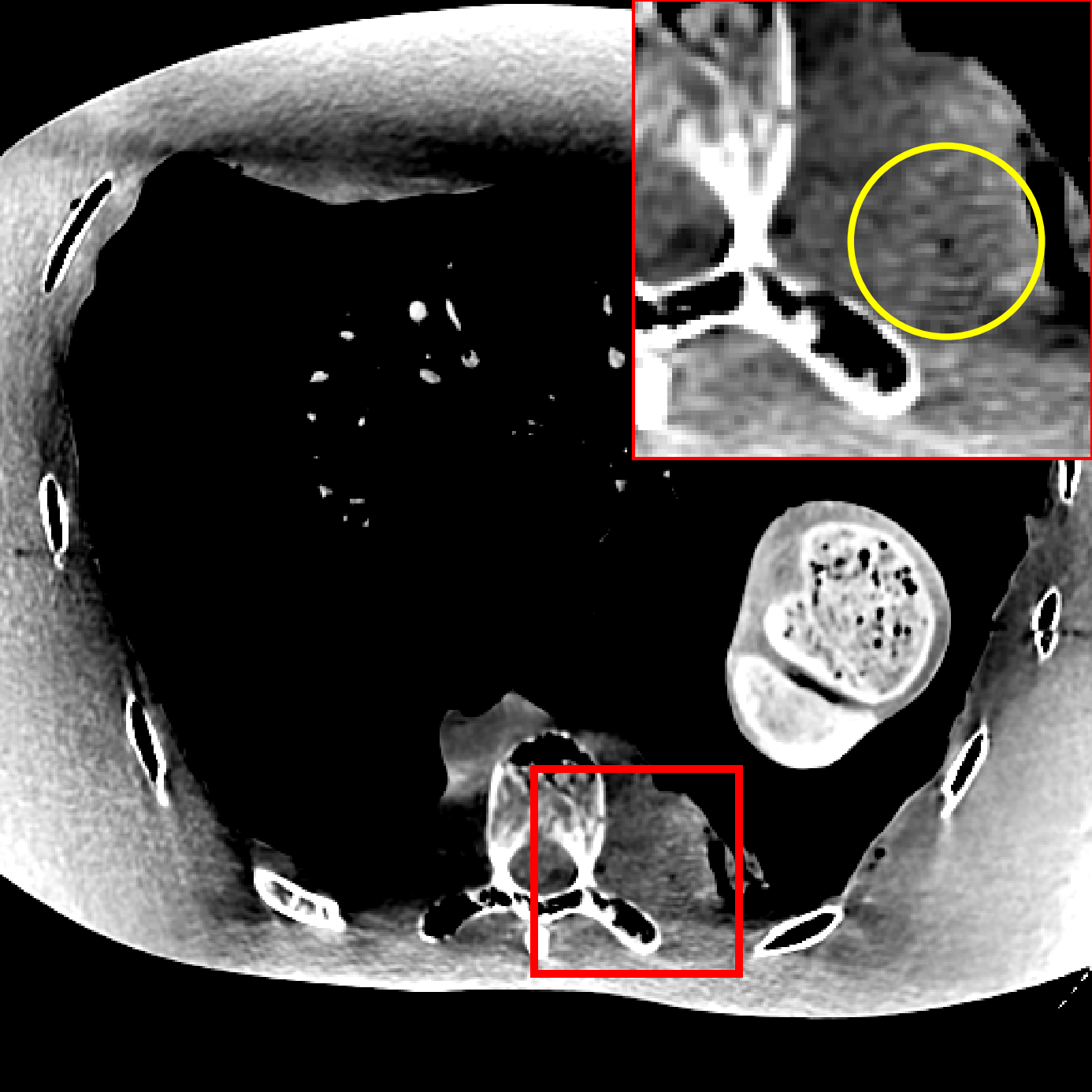}}%
    \subfloat[UNet+NADD]{\includegraphics[width=0.24\textwidth]{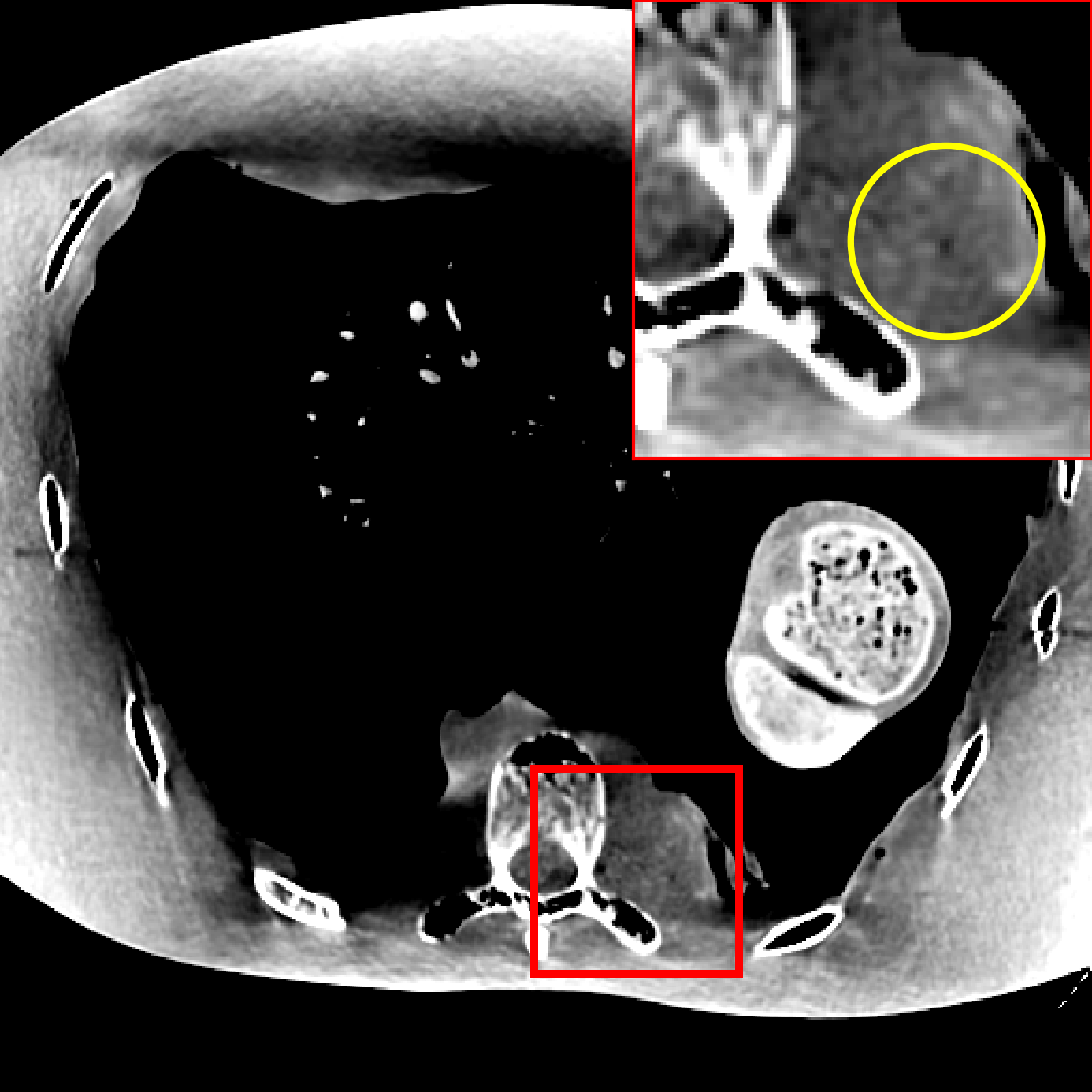}}%
    \subfloat[UNet+COV (Ours)]{\includegraphics[width=0.24\textwidth]{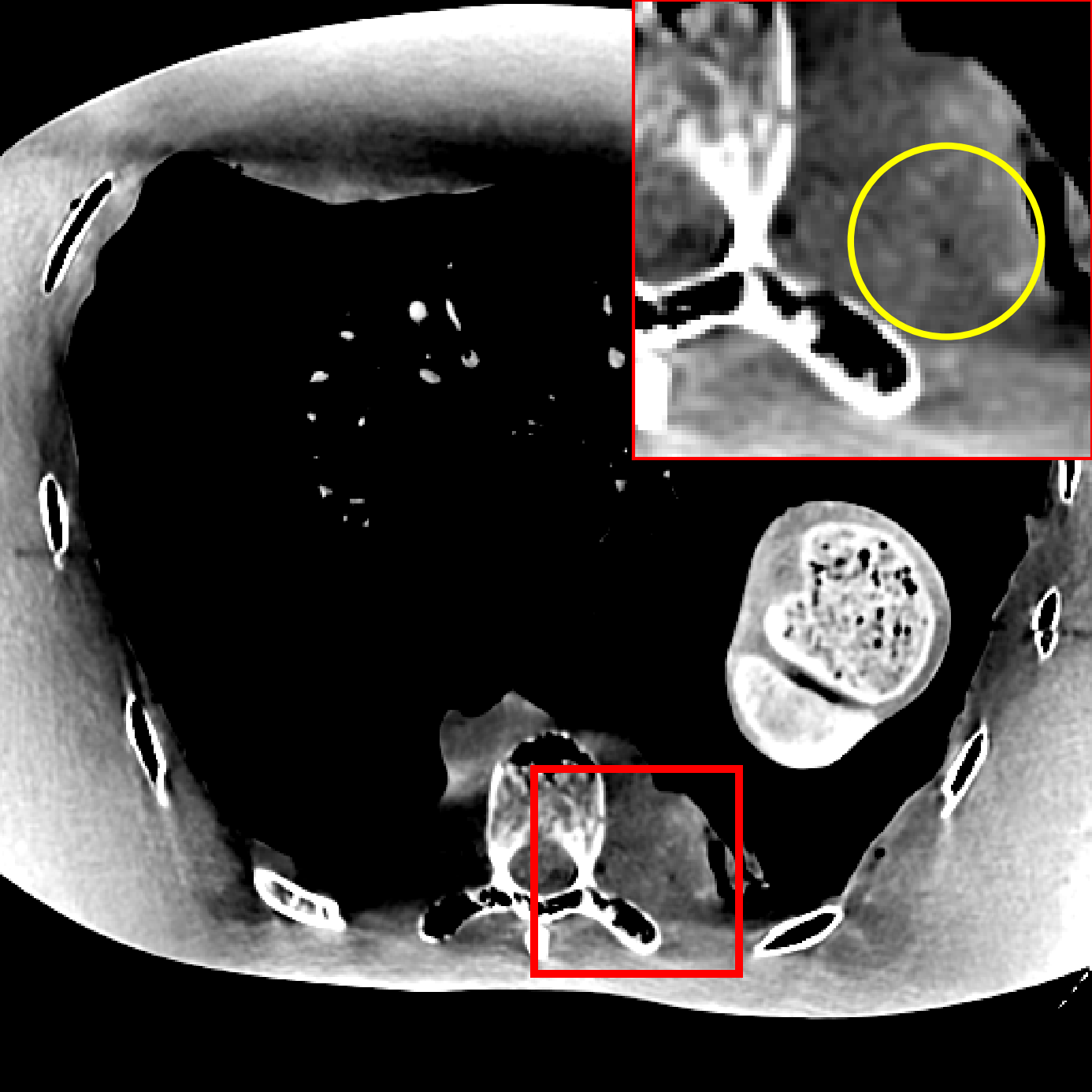}}%
    \subfloat[\textbf{UNet+ENCORE (Ours)}]{\includegraphics[width=0.24\textwidth]{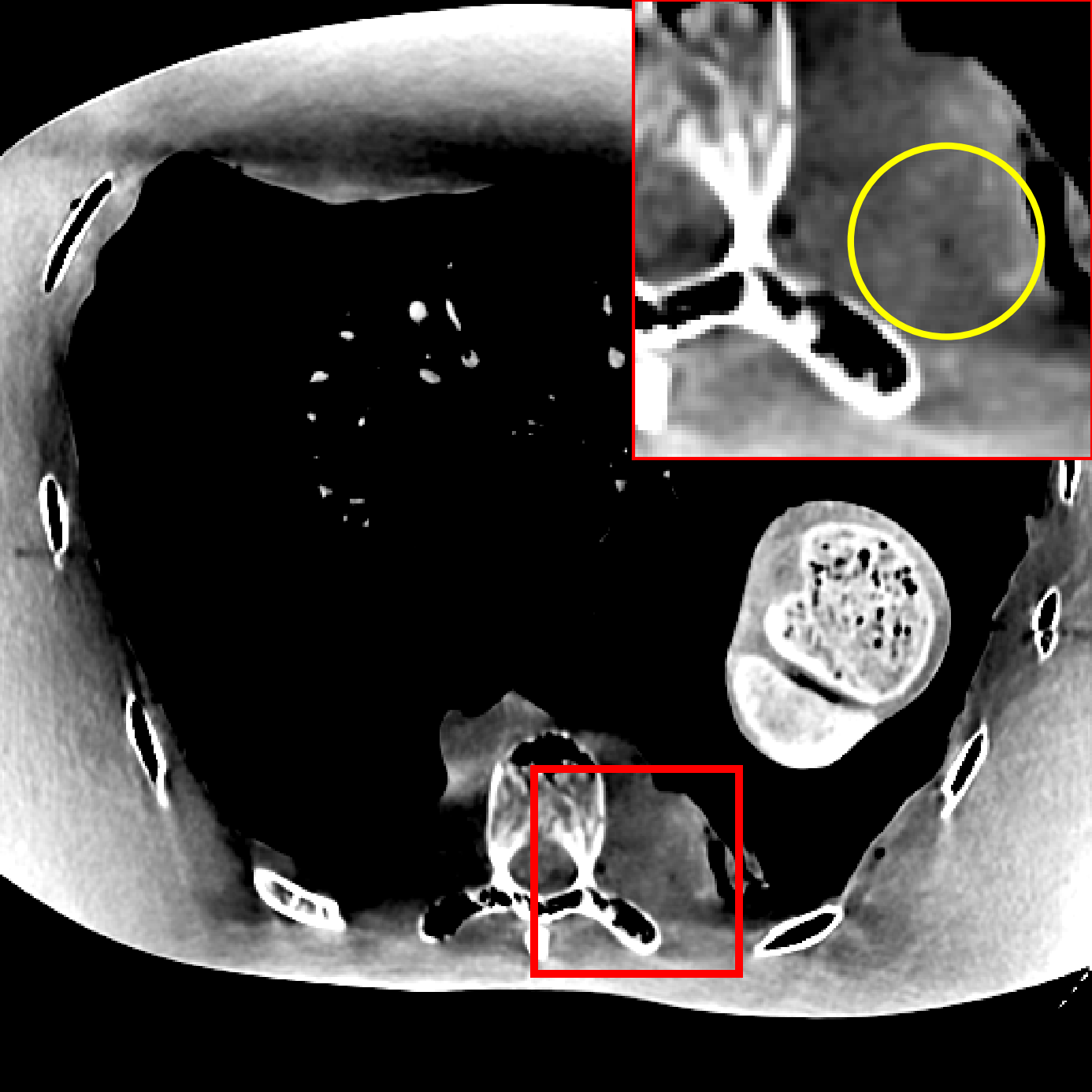}}%
    \caption{Qualitative comparison on the real-world Tabletop dataset with zoomed-in ROI patches (display window: [-200, 40]~HU). UNet variants are evaluated using the Lite configuration. Note that the GT image is unavailable due to dose limits.}
    \label{fig:qualitative_tabletop}
\end{figure}

\section{Results}
\subsection{Qualitative Evaluation}
\cref{fig:qualitative_mayo2016,fig:qualitative_mayo2020,fig:qualitative_tabletop} show qualitative comparisons of each configuration across the Mayo2016, Mayo2020, and Tabletop datasets, respectively.
In \cref{fig:qualitative_mayo2016}, the thick bilateral pelvic bones induce strong horizontal noise correlation and elevated noise power within the central region compared to other slices.
Although these noise patterns are present within the training dataset, models that solely rely on the noisy LDCT input, or those failing to estimate the noise context accurately, struggle to adapt to such spatially-varying noise properties.
Consequently, as shown in the zoomed ROIs (indicated by the yellow circles), alternative configurations suffer from severe detail loss or introduce structural blurring.
In particular, CoreDiff exhibits noticeable residual artifacts due to its limited generalizability to dose levels lower than those encountered during training.
While FoundDiff demonstrates reasonable detail preservation by incorporating anatomical prior knowledge, our proposed +ENCORE jointly leverages anatomical features and explicit noise contexts, restoring fine structures with high fidelity while operating under a substantially lower computational budget. 

To evaluate cross-vendor generalization capability, \cref{fig:qualitative_mayo2020} shows results on the Mayo2020 dataset acquired from GE Healthcare scanners, contrasting with the Mayo2016 dataset acquired from Siemens Healthineers scanners.
Unlike the Siemens dataset which shares the same scanning geometry as the training dataset, the GE dataset is characterized by different geometry parameters, spectrum, and reconstruction kernel, causing the underlying noise context to deviate from the training distribution.
Lacking explicit noise context awareness, Vanilla and diffusion-based CT denoisers struggle to adapt to vendor-specific noise variations, introducing noticeable artifacts as highlighted in the yellow circles of zoomed ROIs. 
In contrast, incorporating noise context information successfully mitigates the resulting distortions, with our +ENCORE achieving effective artifact suppression and enhanced robustness under cross-vendor domain shifts.

Finally, \cref{fig:qualitative_tabletop} displays the evaluation on real-world Tabletop scan data.
This dataset not only differs in scanning geometry from the training setup, but also involves complex physical interactions absent in the simulated datasets.
For instance, residual scatter even after collimation lowers the reconstructed HU values, while pixel crosstalk within the flat-panel detector amplifies the spatial correlation of noise.
Such physical discrepancies from the simulated training distribution serve as a primary cause of denoising performance degradation.
Although the scanned phantom is known to consist of a uniform material without internal textures around the bone structures, both the Vanilla baseline and diffusion-based models exhibit severe artifacts across the surrounding regions due to their lack of noise context awareness.
In contrast, configurations that incorporate noise context information effectively suppress these artifacts, with our +ENCORE demonstrating the best alignment with the expected uniform characteristics of the phantom.

\begin{table}[!t]
    \scriptsize
    \centering
    \caption{Quantitative evaluation on the Mayo2016 dataset across quality metrics (PSNR $\uparrow$, SSIM $\uparrow$, and AUHOC $\downarrow$) and computational efficiency (MACs (G) and latency (ms)). Best results within each network and scale variant are indicated in \textbf{bold}, and \underline{underlined} values denote statistically significant superiority over all configurations.}
    \label{tab:quantitative_mayo2016}
    \vspace*{2ex}
    \setlength{\tabcolsep}{2.5pt}
    \renewcommand{\arraystretch}{1.1}
    \begin{tabular}{c|c|c|c|c|c|c}
        \hline
        Network & Variant & Configuration & $10\%$ & $25\%$ & MACs & Latency \\
        \hline\hline
        LDCT & - & - & 15.47 / 0.3545 / 0.6455 & 19.28 / 0.4548 / 0.5563 & 3.88 & 0.68 \\
        \hline
        \multirow{8}{*}{DnCNN} & \multirow{4}{*}{Lite} & Vanilla & 28.08 / 0.7475 / 0.4776 & 30.58 / 0.8143 / 0.3879 & 13.88 & 2.69 \\
         & & +NADD & 28.36 / 0.7540 / 0.4714 & 30.65 / 0.8151 / 0.3867 & 17.93 & 3.61 \\
         \cdashline{3-7}
         & & \textbf{+COV (Ours)} & 28.57 / 0.7597 / 0.4607 & 30.79 / 0.8186 / 0.3793 & 21.72 & 3.93 \\
         & & \textbf{+ENCORE (Ours)} & \underline{\textbf{29.33}} / \underline{\textbf{0.7766}} / \underline{\textbf{0.4403}} & \underline{\textbf{31.18}} / \underline{\textbf{0.8250}} / \underline{\textbf{0.3634}} & 10.19 & 4.13 \\
        \cline{2-7}
         & \multirow{4}{*}{Base} & Vanilla & 29.44 / 0.7801 / 0.4309 & 31.26 / 0.8244 / 0.3604 & 62.37 & 7.62 \\
         & & +NADD & 29.53 / 0.7822 / 0.4286 & 31.30 / 0.8266 / 0.3588 & 66.42 & 8.55 \\
         \cdashline{3-7}
         & & \textbf{+COV (Ours)} & 29.56 / 0.7833 / 0.4258 & 31.33 / 0.8270 / 0.3568 & 70.21 & 8.90 \\
         & & \textbf{+ENCORE (Ours)} & \underline{\textbf{29.85}} / \underline{\textbf{0.7877}} / \underline{\textbf{0.4165}} & \underline{\textbf{31.46}} / \underline{\textbf{0.8302}} / \underline{\textbf{0.3513}} & 43.17 & 10.14 \\
        \hline
        \multirow{8}{*}{UNet} & \multirow{4}{*}{Lite} & Vanilla & 29.96 / 0.7910 / 0.4104 & 31.58 / 0.8307 / 0.3474 & 18.39 & 4.29 \\
         & & +NADD & 30.05 / 0.7924 / 0.4071 & 31.60 / \underline{\textbf{0.8321}} / 0.3464 & 22.36 & 5.31 \\
         \cdashline{3-7}
         & & \textbf{+COV (Ours)} & 30.10 / 0.7925 / 0.4058 & 31.60 / 0.8299 / 0.3462 & 24.28 & 6.05 \\
         & & \textbf{+ENCORE (Ours)} & \underline{\textbf{30.15}} / \underline{\textbf{0.7943}} / \underline{\textbf{0.4022}} & \underline{\textbf{31.63}} / 0.8306 / \underline{\textbf{0.3444}} & 17.57 & 5.93 \\
        \cline{2-7}
         & \multirow{4}{*}{Base} & Vanilla & 30.01 / 0.7924 / 0.4070 & 31.63 / 0.8319 / 0.3441 & 61.75 & 8.38 \\
          & & +NADD & 30.07 / 0.7928 / 0.4060 & 31.65 / \underline{\textbf{0.8325}} / 0.3440 & 65.79 & 9.53 \\
          \cdashline{3-7}
          & & \textbf{+COV (Ours)} & 30.15 / 0.7942 / 0.4011 & 31.65 / 0.8308 / 0.3438 & 69.47 & 10.32 \\
          & & \textbf{+ENCORE (Ours)} & \underline{\textbf{30.19}} / \underline{\textbf{0.7952}} / \underline{\textbf{0.4002}} & \underline{\textbf{31.69}} / 0.8320 / \underline{\textbf{0.3420}} & 45.05 & 10.24 \\
        \hline
        RED-CNN & - & Vanilla & 29.83 / 0.7889 / 0.4135 & 31.51 / 0.8297 / 0.3489 & 462.28 & 28.18 \\
        \hline
        \multirow{2}{*}{NAFNet} & Lite & Vanilla & 29.91 / 0.7903 / 0.4132 & 31.59 / 0.8316 / 0.3470 & 19.32 & 9.61 \\
        \cline{2-7}
         & Base & Vanilla & 29.92 / 0.7887 / 0.4131 & 31.66 / 0.8327 / 0.3432 & 64.45 & 18.94 \\
        \hline
        \multirow{2}{*}{Uformer} & Lite & Vanilla & 30.04 / 0.7933 / 0.4070 & 31.62 / 0.8322 / 0.3453 & 17.83 & 19.43 \\
        \cline{2-7}
         & Base & Vanilla & 30.07 / 0.7934 / 0.4060 & 31.67 / 0.8328 / 0.3430 & 51.65 & 30.52 \\
        \hline
        \multirow{2}{*}{CoreDiff} & - & Single$^{1}$ & 28.73 / 0.7649 / 0.4531 & 31.03 / 0.8242 / 0.3688 & 1497.46 & 104.90 \\
        \cline{2-7}
        & - & Triplet$^{1, 2}$ & 29.80 / 0.7910 / 0.4204 & 31.75 / 0.8396 / 0.3447 & 1508.25 & 105.92 \\
        \hline
        FoundDiff & - & Vanilla$^{1, 3}$ & 29.98 / 0.7916 / 0.4092 & 31.68 / 0.8320 / 0.3418 & 622.01 & 136.11 \\
        \hline
    \end{tabular}
    \vspace{2pt}
    \begin{minipage}{\linewidth}
        \footnotesize
        $^1$ Trained in a N2C regime with paired supervision, as N2N training is intractable for this architecture.\\
        $^2$ Unlike all other methods that operate on a single slice, it utilizes adjacent slices as multi-channel input, providing richer inter-slice context.\\
        $^3$ Evaluated with an official pre-trained model exposed to diverse dose levels, offering advantages on unseen dose conditions and generalization.
    \end{minipage}
\end{table}

\begin{table}[!t]
    \centering
    \scriptsize
    \caption{Quantitative evaluation on the Mayo2020 dataset across quality metrics (PSNR $\uparrow$, SSIM $\uparrow$, and AUHOC $\downarrow$). Best results within each network and scale variant are indicated in \textbf{bold}, and \underline{underlined} values denote statistically significant superiority over all configurations.}
    \label{tab:quantitative_mayo2020}
    \vspace*{2ex}
    \setlength{\tabcolsep}{4pt}
    \renewcommand{\arraystretch}{1.1}
    \begin{tabular}{c|c|c|c|c}
        \hline
        Network & Variant & Configuration & $10\%$ & $25\%$ \\
        \hline\hline
        LDCT & - & - & 16.96 / 0.4684 / 0.5514 & 20.74 / 0.5677 / 0.4367 \\
        \hline
        \multirow{8}{*}{DnCNN} & \multirow{4}{*}{Lite} & Vanilla & 30.20 / 0.8051 / 0.4537 & 32.32 / 0.8559 / 0.3847 \\
         & & +NADD & 30.40 / 0.8107 / 0.4483 & 32.47 / 0.8586 / 0.3799 \\
         \cdashline{3-5}
         & & \textbf{+COV (Ours)} & 30.57 / 0.8136 / 0.4419 & 32.66 / 0.8606 / 0.3740 \\
         & & \textbf{+ENCORE (Ours)} & \underline{\textbf{31.23}} / \underline{\textbf{0.8298}} / \underline{\textbf{0.4224}} & \underline{\textbf{33.06}} / \underline{\textbf{0.8676}} / \underline{\textbf{0.3596}} \\
        \cline{2-5}
         & \multirow{4}{*}{Base} & Vanilla & 31.42 / 0.8340 / 0.4161 & 33.12 / 0.8681 / 0.3583 \\
         & & +NADD & 31.48 / 0.8344 / 0.4134 & 33.20 / 0.8693 / 0.3557 \\
         \cdashline{3-5}
         & & \textbf{+COV (Ours)} & 31.53 / 0.8357 / 0.4101 & 33.25 / 0.8701 / 0.3532 \\
         & & \textbf{+ENCORE (Ours)} & \underline{\textbf{31.80}} / \underline{\textbf{0.8408}} / \underline{\textbf{0.4021}} & \underline{\textbf{33.41}} / \underline{\textbf{0.8729}} / \underline{\textbf{0.3480}} \\
        \hline
        \multirow{8}{*}{UNet} & \multirow{4}{*}{Lite} & Vanilla & 31.87 / 0.8419 / 0.3964 & 33.49 / 0.8725 / 0.3451 \\
         & & +NADD & 31.99 / 0.8445 / 0.3940 & 33.55 / 0.8744 / 0.3429 \\
         \cdashline{3-5}
         & & \textbf{+COV (Ours)} & 32.04 / 0.8449 / 0.3932 & 33.57 / 0.8748 / 0.3430 \\
         & & \textbf{+ENCORE (Ours)} & \underline{\textbf{32.06}} / \underline{\textbf{0.8459}} / \underline{\textbf{0.3925}} & \underline{\textbf{33.60}} / \textbf{0.8753} / \underline{\textbf{0.3422}} \\
        \cline{2-5}
         & \multirow{4}{*}{Base} & Vanilla & 32.01 / 0.8443 / 0.3928 & 33.58 / 0.8743 / 0.3420 \\
          & & +NADD & 32.03 / 0.8451 / 0.3927 & 33.61 / 0.8751 / 0.3410 \\
          \cdashline{3-5}
          & & \textbf{+COV (Ours)} & 32.05 / 0.8453 / 0.3912 & 33.63 / 0.8757 / 0.3397 \\
          & & \textbf{+ENCORE (Ours)} & \textbf{32.07} / \underline{\textbf{0.8462}} / \textbf{0.3908} & \underline{\textbf{33.66}} / \underline{\textbf{0.8763}} / \textbf{0.3392} \\
        \hline
        RED-CNN & - & Vanilla & 31.84 / 0.8406 / 0.3986 & 33.47 / 0.8725 / 0.3452 \\
        \hline
        \multirow{2}{*}{NAFNet} & Lite & Vanilla & 31.90 / 0.8419 / 0.3971 & 33.53 / 0.8736 / 0.3448 \\
        \cline{2-5}
         & Base & Vanilla & 31.88 / 0.8416 / 0.3959 & 33.58 / 0.8743 / 0.3418 \\
        \hline
        \multirow{2}{*}{Uformer} & Lite & Vanilla & 32.06 / 0.8460 / 0.3921 & 33.61 / 0.8748 / 0.3418 \\
        \cline{2-5}
         & Base & Vanilla & 32.08 / 0.8462 / 0.3906 & 33.65 / 0.8755 / 0.3397 \\
        \hline
        \multirow{2}{*}{CoreDiff} & - & Single$^{1}$ & 30.62 / 0.8126 / 0.4340 & 32.78 / 0.8616 / 0.3659 \\
        \cline{2-5}
        & - & Triplet$^{1, 2}$ & 30.90 / 0.8270 / 0.4206 & 32.81 / 0.8704 / 0.3484 \\
        \hline
        FoundDiff & - & Vanilla$^{1, 3}$ & 31.85 / 0.8406 / 0.3963 & 33.59 / 0.8741 / 0.3410 \\
        \hline
    \end{tabular}
    \vspace{2pt}
    \begin{minipage}{\linewidth}
        \footnotesize
        $^1$ Trained in a N2C regime with paired supervision, as N2N training is intractable for this architecture.\\
        $^2$ Unlike all other methods that operate on a single slice, it utilizes adjacent slices as multi-channel input, providing richer inter-slice context.\\
        $^3$ Evaluated with an official pre-trained model exposed to diverse dose levels, offering an advantage on unseen dose conditions.
    \end{minipage}
\end{table}

\subsection{Quantitative Evaluation}
\cref{tab:quantitative_mayo2016,tab:quantitative_mayo2020} present the quantitative results for each configuration on the Mayo2016 and Mayo2020 datasets, respectively.
In terms of theoretical computational complexity, incorporating our proposed ENCORE framework yields a substantial reduction in MACs compared to the other configurations. 
Additionally, the actual inference latency remains comparable to that of the +COV variant, indicating that the proposed structure does not impose excessive latency overhead. 
In contrast, attention-driven architectures (e.g., NAFNet \cite{nafnet} and Uformer \cite{uformer}) incur substantial latency overheads under comparable MACs, even when optimized via kernel fusion.
A detailed discussion on the hardware-level translation gap is provided in \cref{sec:discussion}.

Regarding the image quality metrics on the Mayo2016 dataset (\cref{tab:quantitative_mayo2016}), across both UNet \cite{unet} and DnCNN \cite{dncnn} backbones, incorporating noise context profiles consistently enhances performance. 
Specifically, +COV, which incorporates our proposed autocovariance estimation as a pre-processing step, outperforms +NADD in most cases.
This tendency indicates that the autocovariance map provides more effective and model-assimilable noise context information.
Moreover, thanks to the FlyingConv structure that unlocks the latent potential of the autocovariance map, the integration of our proposed ENCORE further improves denoising quality, outperforming the +COV configuration.
In particular, conditioning on the local noise context helps mitigate correlated noise artifacts, yielding overall quality improvements alongside favorable AUHOC trends.
This denoising quality gain is pronounced in Lite variants as well as the $10\%$ ultra-low-dose regime unseen during training.
On the other hand, the Vanilla and +NADD variants exhibit higher SSIM scores than ENCORE in the UNet on the 25\% dose level.
This lower SSIM score under the default $d_{\text{target}} = \infty$ setting can be dynamically resolved by adjusting $d_{\text{target}}$ during inference. 
As detailed in \cref{sec:zero_shot}, such inference-time tuning not only resolves this gap to surpass both baselines, but also further amplifies ENCORE's advantage in cases where it already achieves higher SSIM.
In comparison with other advanced baselines, attention-driven models (e.g., NAFNet \cite{nafnet} and Uformer \cite{uformer}) suffer from latency overheads while offering suboptimal denoising fidelity compared to ENCORE.
For diffusion-based methods, although CoreDiff \cite{corediff} in its Triplet configuration achieves high SSIM via 3-slice adjacent input, it exhibits an increased susceptibility to structurally inconsistent artifacts (indicated by higher AUHOC) and limited robustness under unseen dose levels (which drops further in the single-slice setting).
Meanwhile, FoundDiff \cite{founddiff} demonstrates stable robustness thanks to its foundation pre-training on diverse anatomical and dose distributions, yet it falls short of +ENCORE across all cases despite its substantial computational overhead.
Such performance limitations in both diffusion models may partly stem from geometric discrepancies.
While their original studies were evaluated under uniform voxel spacing, our pipeline preserves patient-specific pixel spacing to reflect real-world clinical acquisitions, which introduces realistic scale variations that can challenge diffusion-based models.

Next, the Mayo2020 dataset (\cref{tab:quantitative_mayo2020}) evaluates cross-vendor generalization on a GE Healthcare scanner, whose distinct scanning geometry and reconstruction alters spatial noise characteristics.
Similarly, +ENCORE consistently improves image quality over the conventional configurations (Vanilla and +NADD), confirming that explicit noise context conditioning robustly handles geometric domain shifts.
Notably, CoreDiff-Triplet, which previously demonstrated the highest PSNR and SSIM scores under in-distribution training conditions, substantially lags behind +ENCORE across all cases, revealing its vulnerability to noise distribution shifts caused by geometric discrepancies.
Likewise, although FoundDiff holds an intrinsic advantage from being pre-trained on diverse dose levels including the Mayo2020 cohort, its overall denoising quality consistently falls short of our +ENCORE across all metrics.

\subsection{Ablation Studies}
\subsubsection{Skewness-corrected noise modeling}
We evaluate the impact of replacing the Gaussian approximated term $\mathcal{N}(0, P_\text{ND})$ in \cref{eq:n2n_generation} with the skewness-corrected term $\mathcal{W}$ derived from the Cornish-Fisher expansion to generate $P_\text{sLD}$ for N2N training.
As shown in the upper region of \cref{tab:noise_ablation}, the Gaussian approximation (which has zero skewness) is valid only as the Poisson noise skewness approaches zero at high photon counts, but fails at low photon counts. 
In contrast, our formulation closely tracks the target Poisson skewness across all photon counts.
This improved statistical agreement translates to enhanced denoising performance, as shown in the lower region of \cref{tab:noise_ablation}.
Here, test slices are categorized based on anatomical structures: slices containing dense and thick bone structures (e.g., pelvic and femur bones) are grouped into the high-attenuation subset, while the remaining slices form the low-attenuation subset. 
The gain in PSNR remains marginal across all regions because global pixel-wise $L_2$ error is relatively insensitive to higher-order noise statistics such as skewness.
Conversely, SSIM steadily improves in high-attenuation slices, confirming that skewness correction mitigates non-Gaussian distortions under photon starvation conditions.
In low-attenuation regions, where photon counts are sufficiently high for the Gaussian approximation to remain valid, the benefit of skewness correction becomes less pronounced.

\begin{table}
\centering
\caption{Theoretical skewness and quality metrics (PSNR / SSIM) on low/high-attenuation slices evaluated with the Base UNet+ENCORE on the Mayo2016 dataset.}
\label{tab:noise_ablation}
\vspace*{2ex}
\begin{tabular}{lcccc}
\hline
Photon Count & 25 & 250 & 2,500 & 25,000 \\
\hline
Poisson (Target) & 0.1501 & 0.0579 & 0.0187 & 0.0059 \\
\textbf{Ours} & 0.1499 & 0.0579 & 0.0187 & 0.0059 \\
\hline
\hline
Denoising Method & \multicolumn{2}{c}{Low-Atten.} & \multicolumn{2}{c}{High-Atten.} \\
\hline
Gaussian & \multicolumn{2}{c}{31.74 / 0.8344} & \multicolumn{2}{c}{31.38 / 0.8186} \\
\textbf{Ours} & \multicolumn{2}{c}{\textbf{31.74} / \textbf{0.8347}} & \multicolumn{2}{c}{\textbf{31.39} / \textbf{0.8203}} \\
\hline
\end{tabular}
\end{table}

\subsubsection{Zero-shot denoising evaluation}
\label{sec:zero_shot}
\begin{figure}
    \centering
    \subfloat[]{\label{fig:zero_shot_metrics}%
        \includegraphics[width=0.8\linewidth]{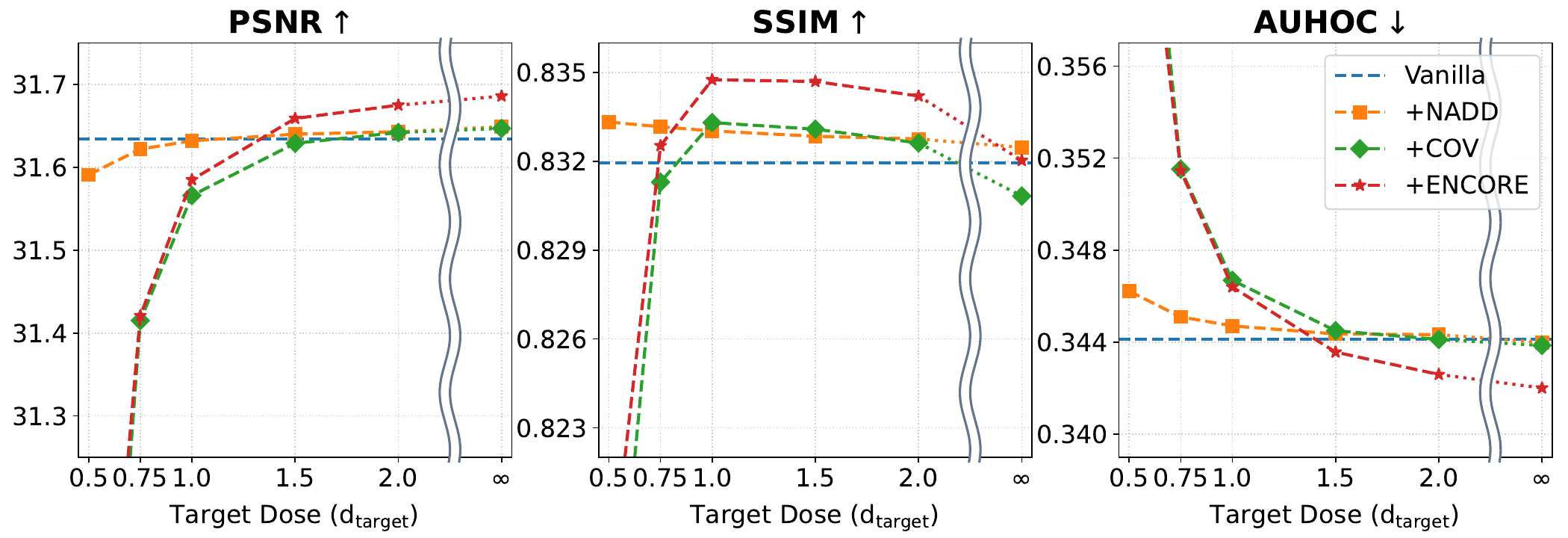}%
    }
    \par
    \subfloat[]{\label{fig:zero_shot_visual}%
        \begin{minipage}{\linewidth}
            \centering
            \begin{minipage}[t]{0.2\linewidth}
                \centering
                \includegraphics[width=\linewidth]{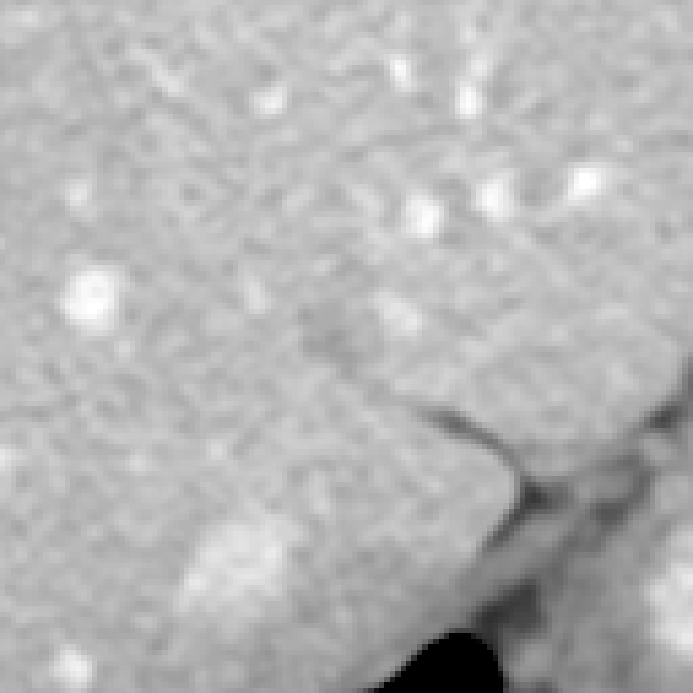}\\[-1.0ex]
                {\footnotesize $d_{\text{target}} = \infty$}
            \end{minipage}%
            \begin{minipage}[t]{0.2\linewidth}
                \centering
                \includegraphics[width=\linewidth]{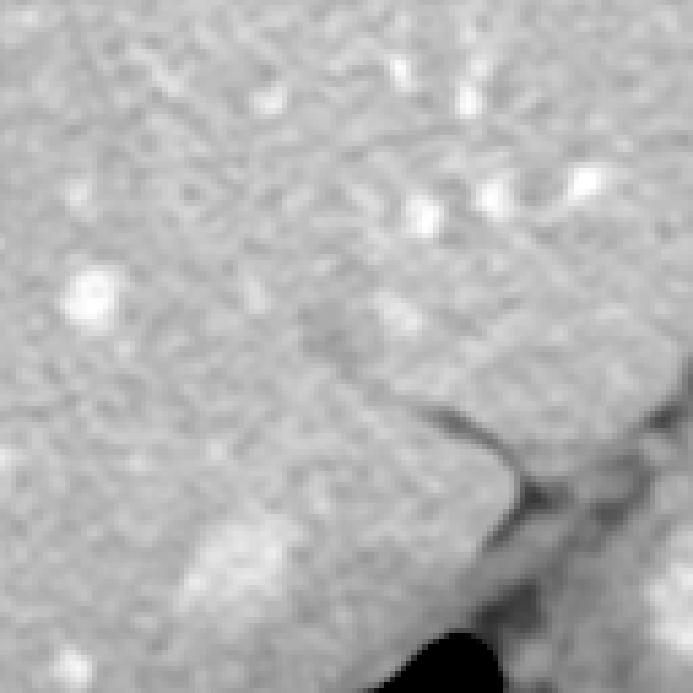}\\[-1.0ex]
                {\footnotesize $d_{\text{target}} = 2.0$}
            \end{minipage}%
            \begin{minipage}[t]{0.2\linewidth}
                \centering
                \includegraphics[width=\linewidth]{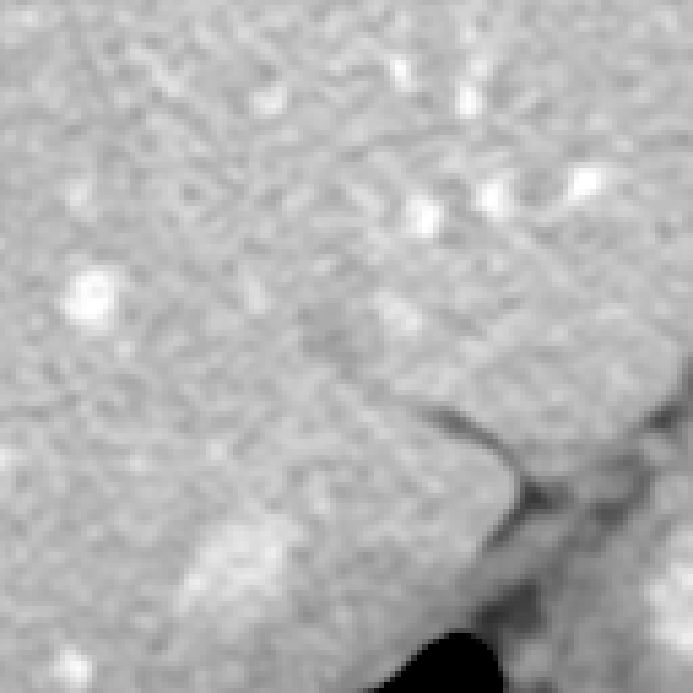}\\[-1.0ex]
                {\footnotesize $d_{\text{target}} = 1.5$}
            \end{minipage}%
            \begin{minipage}[t]{0.2\linewidth}
                \centering
                \includegraphics[width=\linewidth]{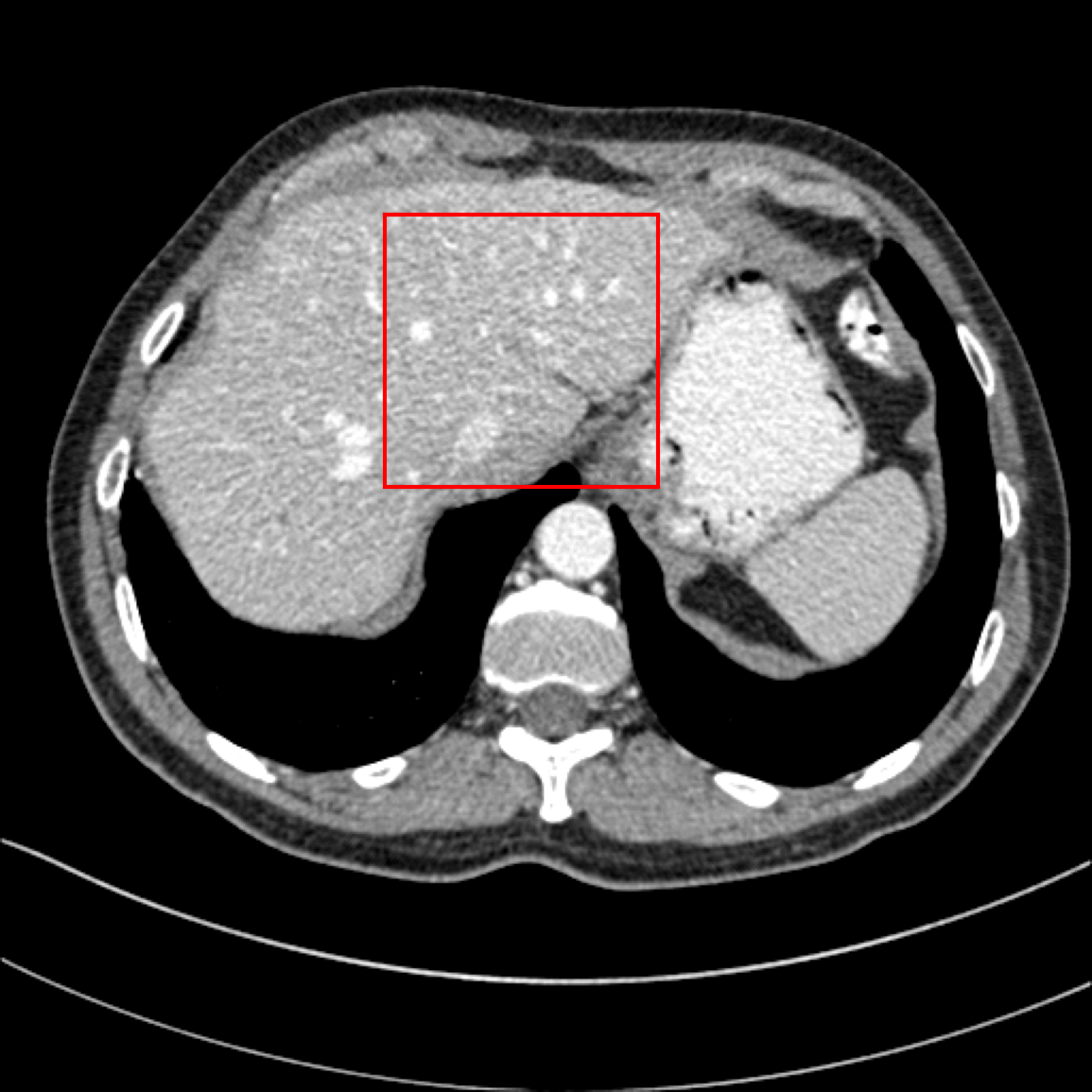}\\[-1.0ex]
            \end{minipage} \\ \vspace{0.5mm}
            \begin{minipage}[t]{0.2\linewidth}
                \centering
                \includegraphics[width=\linewidth]{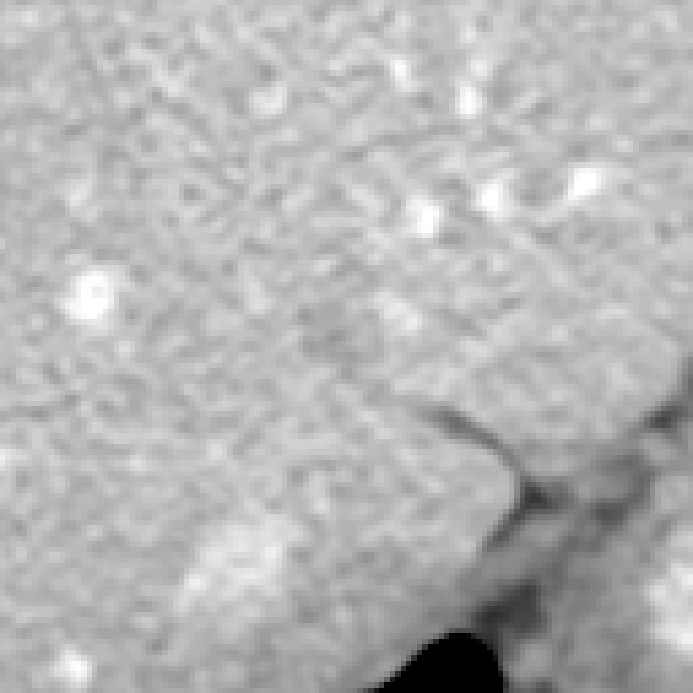}\\[-1.0ex]
                {\footnotesize $d_{\text{target}} = 1.0$}
            \end{minipage}%
            \begin{minipage}[t]{0.2\linewidth}
                \centering
                \includegraphics[width=\linewidth]{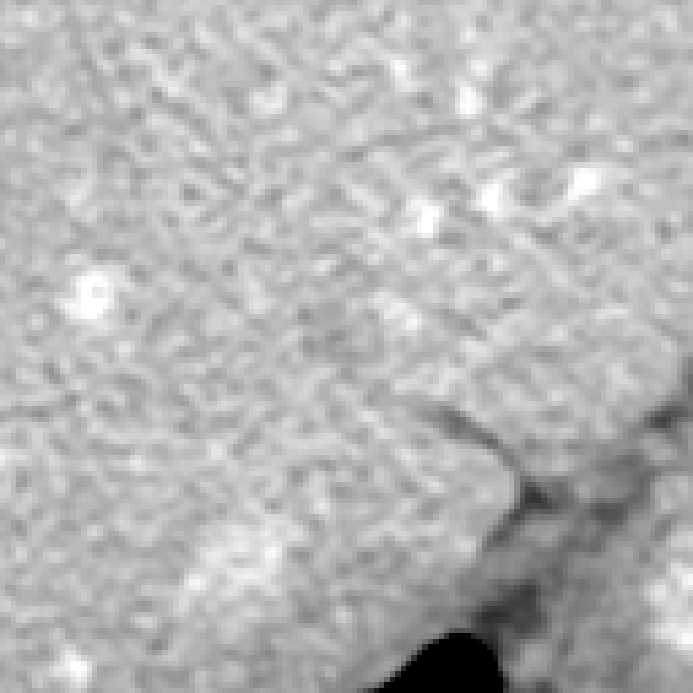}\\[-1.0ex]
                {\footnotesize $d_{\text{target}} = 0.75$}
            \end{minipage}%
            \begin{minipage}[t]{0.2\linewidth}
                \centering
                \includegraphics[width=\linewidth]{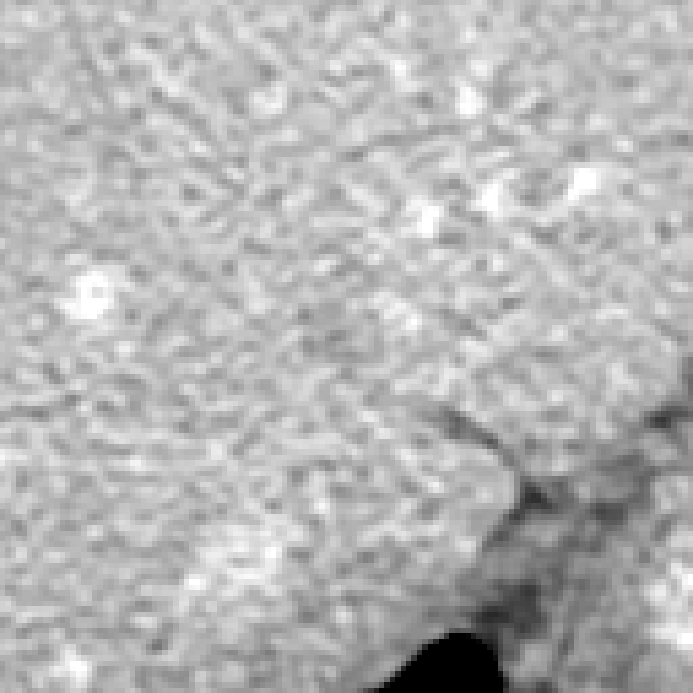}\\[-1.0ex]
                {\footnotesize $d_{\text{target}} = 0.5$}
            \end{minipage}%
            \begin{minipage}[t]{0.2\linewidth}
                \centering
                \includegraphics[width=\linewidth]{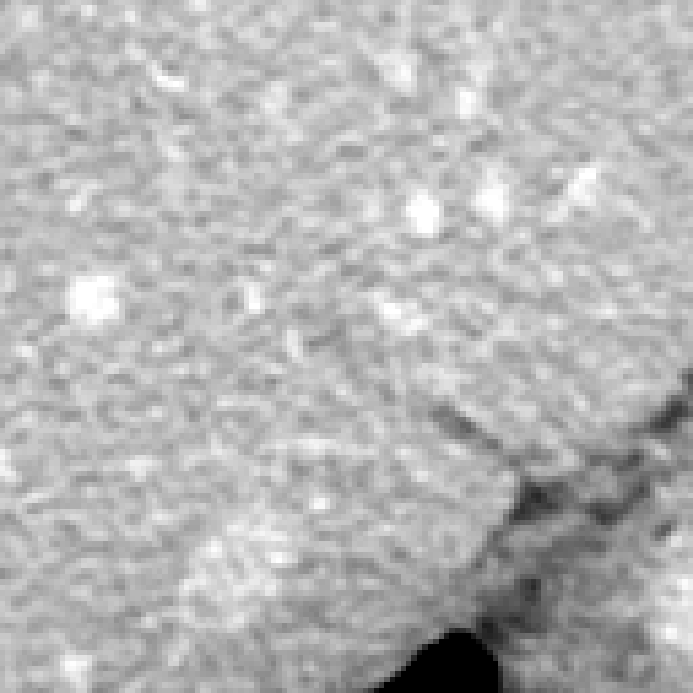}\\[-1.0ex]
                {\footnotesize GT}
            \end{minipage}
        \end{minipage}%
    }
    \caption{Result for sweeping $d_{\text{target}}$ values, evaluated on the Mayo2016 dataset at a 25\% dose level with the Base UNet+ENCORE. (a) Image quality metrics (b) ROI patches.}
    \label{fig:zero_shot_ablation}
\end{figure}

Our proposed ENCORE framework not only boosts denoising quality, but also grants architectural flexibility, enabling zero-shot adjustments tailored to clinical preferences.
As shown in \cref{fig:zero_shot_metrics}, sweeping the $d_{\text{target}}$ value reveals a clear trade-off between metrics.
In detail, while targeting noise-free images ($d_{\text{target}}=\infty$) yields the best PSNR and AUHOC, SSIM improves when targeting noise-containing images ($d_{\text{target}}<\infty$) because the residual noise can prevent over-smoothing.
This enables dynamic modulation of output texture during inference without any additional training or post-processing.
Although the GT image is treated as noise-free in our experiment, the clinical NDCT references still contain a small amount of noise;
thus, setting $d_{\text{target}}=\infty$ shows minimum pixel-level errors but leads to over-smoothing. 
On the other hand, setting $d_{\text{target}} \in [0.75, 1.0]$ yields visually favorable results by matching the realistic texture and noise level of clinical scans as demonstrated in \cref{fig:zero_shot_visual}.
Furthermore, when the SSIM of all variants is aligned to a similar level, ENCORE exhibits superior quality in terms of PSNR and AUHOC, minimizing the intrinsic trade-off.
This interactive zero-shot capability also resolves the marginal SSIM degradation observed under the default $d_{\text{target}} = \infty$ setting in \cref{tab:quantitative_mayo2016,tab:quantitative_mayo2020}.

\subsubsection{Noise context estimation configurations} 
\cref{tab:ablation_noisecontext} shows the results of different noise context configurations. 
Regarding the window size, using $w=5$ yields the best performance, outperforming the baseline ($w=1$) which strictly adheres to the original autocovariance definition.
This improvement is consistent with the prior finding that CT noise distribution can be approximated as a stationary process within a small ROI \cite{baek2010noise}. 
In terms of patch size, $p=5$ yields the best results, as it spans an area that is neither too localized nor overly broad for $3 \times 3$ convolution kernels.
Finally, for the number of noise augmentations $N$, a larger value theoretically improves the autocovariance estimation accuracy.
However, it requires additional FBP executions and increases the computational overhead.
Since no noticeable performance gain is observed, setting $N=1$ is sufficient to capture the noise context.
\begin{table}
    \centering
    \caption{Ablation study on the noise context estimation parameters on the Mayo2016 dataset (25\% dose level) with the Base UNet+ENCORE, where latency excludes model inference.}
    \label{tab:ablation_noisecontext}
    \vspace*{2ex}
    \renewcommand{\arraystretch}{1.1}
    \setlength{\tabcolsep}{8pt}
    \footnotesize
    \begin{tabular}{ccc|cc}
        \hline
        $w$ & $p$ & $N$ & PSNR & Latency (ms) \\
        \hline\hline
        1  & 5 & 1 & 31.66 & 1.58 \\
        \textbf{5} & \textbf{5} & \textbf{1} & \textbf{31.69} & \textbf{1.68} \\
        9 & 5 & 1 & 31.67 & 1.80 \\
        \hline
        5 & 3 & 1 & 31.65 & 1.57 \\
        5 & 7 & 1 & 31.66 & 2.10 \\
        \hline
        5 & 5 & 3 & 31.68 & 3.25 \\
        5 & 5 & 10 & 31.66 & 8.88 \\
        \hline
    \end{tabular}
\end{table}

\subsubsection{FlyingConv module configurations}
As previously mentioned, increasing the group size $g$ reduces the memory access overhead during the FlyingConv operations, substantially lowering latency.
Moreover, moderately increasing $g$ enhances training stability.
However, an excessively large $g$ restricts the independence of the kernel weights, limiting the model's representational capacity.
This trade-off is clearly demonstrated in \cref{tab:ablation_groupsize}, where $g=2$ achieves the optimal balance between computational efficiency and denoising performance. 
Hence, we adopt $g=2$ as the default configuration throughout this study.

\begin{table}
    \centering
    \caption{Ablation study on group size $g$ in FlyingConv, evaluated on Mayo2016 (25\% dose) using Base UNet+ENCORE. Latency covers model inference only.}
    \label{tab:ablation_groupsize}
    \renewcommand{\arraystretch}{1.1}
    \setlength{\tabcolsep}{4pt}
    \footnotesize
    \begin{tabular}{c|c|ccc}
        \hline
        $g$ & PSNR & MACs (G) & Latency (ms) & Peak Memory (MB) \\
        \hline\hline
        1 & 31.67 & 38.23 & 8.97 & 379.64 \\
        \textbf{2} & \textbf{31.69} & \textbf{37.10} & \textbf{8.56} & \textbf{370.64} \\
        4 & 31.67 & 36.54 & 8.49 & 366.14 \\
        8 & 31.66 & 36.26 & 8.42 & 363.89 \\
        \hline
    \end{tabular}
\end{table}

\subsection{Benchmark for customized kernels}
\begin{figure}
    \centering
    \includegraphics[width=0.8\linewidth]{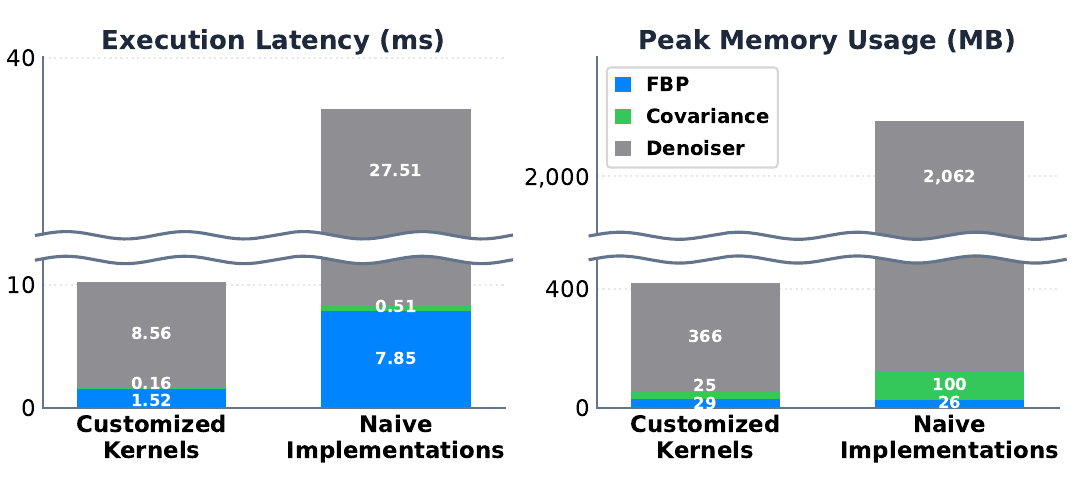}
    \caption{Benchmark comparison between customized CUDA kernels and the naive implementations for the Base UNet+ENCORE, measured at a batch size 16 and normalized per single axial slice.}
    \label{fig:benchmark_kernels}
\end{figure}

\cref{fig:benchmark_kernels} presents the computational benchmark results, comparing efficiency gains achieved by our customized CUDA kernels.
Here, the unoptimized baselines correspond to naive PyTorch implementations for autocovariance estimation and FlyingConv, and vanilla LEAP for FBP reconstruction.
In terms of execution latency, our customized CUDA kernels provide a remarkable throughput boost; the FBP module achieves a $4.8\times$ acceleration compared to LEAP-based FBP, while the covariance estimation and FlyingConv-based denoiser modules deliver $2.9\times$ and $3.2\times$ speedups over their naive PyTorch baselines, respectively.
Beyond latency improvements, our customized kernels reduce the memory footprints of the covariance estimation and denoiser modules by $4.0\times$ and $5.6\times$, respectively.

\section{Discussion}
\label{sec:discussion}
While our approach generally exhibits superior performance compared to other configurations, it is worth discussing why the Vanilla baseline or +NADD occasionally achieves higher SSIM values within \cref{tab:quantitative_mayo2016,tab:quantitative_mayo2020}.
Specifically, under the default setting ($d_{\text{target}} = \infty$), ENCORE achieves higher PSNR by minimizing pixel-wise $L_2$ errors, but can exhibit lower SSIM scores on slices dominated by homogeneous anatomical structures such as the liver.
This phenomenon stems from the aggressive suppression of residual noise:
because clinical reference scans contain slight residual noise, completely removing it can marginally reduce SSIM in homogeneous regions.
Rather than a structural limitation, our framework addresses this classic fidelity-texture dilemma by offering $d_{\text{target}}$ as a zero-shot parameter to preserve preferred clinical textures without re-training.
This capability is particularly beneficial in self-supervised learning environments such as N2N. 
Since N2N-based training lacks clean GT target images, the application of conventional perceptual loss functions (e.g., VGG loss \cite{johnson2016perceptual}) is fundamentally restricted.
Consequently, balancing noise reduction and texture preservation has been a persistent challenge. 
Our framework addresses this limitation by enabling zero-shot control over the denoising texture, providing a level of flexibility that is difficult to achieve solely through loss function modifications.

Regarding computational efficiency, our proposed ENCORE achieves a substantial reduction in MACs across all architectures as shown in \cref{tab:quantitative_mayo2016}. 
Importantly, this massive computational savings is realized while maintaining highly competitive inference latency.
Although dynamic weight modulation via FlyingConv introduces memory access overhead that can partially constrain arithmetic speedups, our customized kernels successfully suppress this overhead, ensuring that ENCORE achieves superior quality without incurring latency penalties.
Furthermore, implementing dedicated backend optimizations such as restructuring the memory access patterns of the FlyingConv module holds promise for unlocking additional efficiency margins of our ENCORE framework.

Regarding practical applicability, our ENCORE was trained exclusively on simulated datasets.
Nevertheless, it maintains favorable performance on the real-world Tabletop dataset, suggesting robustness under physical imaging conditions.
This success highlights the strong generalization capability of ENCORE, demonstrating that the learned noise context representations remain effective even under unmodeled physical factors, such as scatter.
Building on this foundation, future work incorporating comprehensive physical modeling into the noise context estimation framework can further unlock performance enhancements in real-world clinical imaging.

\section{Conclusion}

In this study, we propose an ENCORE framework that integrates noise context into the denoising pipeline.
We perform a full-stack optimization across the entire imaging chain, from image reconstruction to the denoising model, facilitating practical deployment.
Overall, our findings demonstrate that explicit noise context conditioning improves performance with high robustness, while enabling flexible control of output image texture.

\section*{Acknowledgements}
\addcontentsline{toc}{section}{\numberline{}Acknowledgements}
This work was partially supported by the National Research Foundation of Korea (NRF) grant funded by the Korea government through the Ministry of Science and ICT (MSIT) (RS-2023-00240135, RS-2025-00553670), and Institute of Information \& Communications Technology Planning \& Evaluation (IITP) grant funded by MSIT (RS-2020-II201361, Artificial Intelligence Graduate School Program (Yonsei University)).

\section*{Declaration of Generative AI in the Writing Process}
During the preparation of this manuscript, the authors used Google Gemini in order to improve language grammar.
After using this tool, the authors reviewed and edited the content as needed and take full responsibility for the content of the published article.

\section*{Appendix: Noise Synthesis via Cornish-Fisher Expansion}
\addcontentsline{toc}{section}{\numberline{}Appendix}
\label[appendix]{appendix}
While the Poisson quantum noise term can be modeled using a Gaussian approximation, it fails to capture the physical skewness of the Poisson distribution under the low photon counts.
To emulate the noise characteristics more accurately, we construct a noise distribution $\mathcal{W}$ that satisfies both the target variance $V_{\text{target}}$ and skewness $S_{\text{target}}$.
Using the Cornish-Fisher expansion \cite{cornish1938moments}, the synthesized quantum noise $\mathcal{W}$ is defined as follows:
\begin{equation}
    \mathcal{W}(0, V_{\text{target}}) = \sqrt{V_{\text{target}}} (\alpha \mathcal{N}(0,1) + \beta (\mathcal{N}(0,1)^2 - 1)),
\end{equation}
where $\beta$ controls the skewness, and $\alpha = \sqrt{\max(1 - 2\beta^2, 0)}$ ensures $\operatorname{Var}(\mathcal{W}(0, V_{\text{target}})) = V_{\text{target}}$.
By equating the third central moment of $\mathcal{W}(0, V_{\text{target}})$ to the target skewness $S_{\text{target}}$, $\beta$ is derived as:
\begin{equation}
\begin{aligned}
    &E[\mathcal{W}(0, V_{\text{target}})^3] = V_{\text{target}}^{3/2} (6\alpha^2\beta + 8\beta^3) \\
    &\approx 6\beta V_{\text{target}}^{3/2} = S_{\text{target}} \implies \beta \approx \frac{S_{\text{target}}}{6 V_{\text{target}}^{3/2}},
\end{aligned}
\label{eq:beta_approx}
\end{equation}
where we assume that the $\beta$ value is small.
Then, the synthesized noise distribution $\mathcal{W}$ is defined by substituting $V_{\text{target}}$ and $S_{\text{target}}$ values according to the desired imaging conditions.

\subsection*{Noise2Noise Pair Generation for Training}
\label{subsec:n2n_generation}
In order to generate N2N pair data, the variance and skewness of the noise to be added should be calculated, excluding the variance and skewness already contained in $P_{\text{ND}}$.
The variance and skewness to be added are defined as follows:
\begin{equation}
\begin{aligned}
    &V_{\text{target}} = d(1-d) P_{\text{ND}} \\
    &S_{\text{target}} = d(1-d^2) P_{\text{ND}}.
\end{aligned}
\end{equation}
By substituting $V_{\text{target}}$ and $S_{\text{target}}$ values into \cref{eq:beta_approx}, we obtain $\beta \approx \frac{1+d}{6 \sqrt{d(1-d) P_{\text{ND}}}}$.
The synthesized quantum noise $\mathcal{W}$, derived from the calculated $\alpha$ and $\beta$ values, can replace the term $\mathcal{N}(0, P_{\text{ND}})$ in \cref{eq:n2n_generation}.
This allows the synthesized noise to reflect real-world physics compared to the conventional Gaussian approximation. 

\subsection*{Noise Map Generation for Noise Augmentation}
\label{subsec:noise_augmentation}
To synthesize lower-dose data for noise augmentation, additional noise $\mathcal{W}$ is injected into $P_{\text{LD}}$ to accurately emulate actual low-dose noise statistics:
In this case, the $V_{\text{target}}$ and $S_{\text{target}}$ are defined as follows:
\begin{equation}
    V_{\text{target}} = P_{\text{LD}}, S_{\text{target}} = P_{\text{LD}}, \beta \approx \frac{1}{6 \sqrt{P_{\text{LD}}}}.
\end{equation}
Similar to the previous section, $\mathcal{W}$ is estimated using the calculated $\alpha$ and $\beta$ values, and is utilized to replace $\mathcal{N}(0, P_{\text{LD}})$ in \cref{eq:noise_augmentation} to improve the noise augmentation process.

\section*{References}
\addcontentsline{toc}{section}{\numberline{}References}
\vspace*{-10mm}
\bibliographystyle{medphy}    
\bibliography{report}      

\end{document}